\documentclass[twocolumn]{article}
\usepackage[left=1.5cm, right=1.5cm, top=2cm, bottom = 2cm]{geometry}
\usepackage{graphicx}
\usepackage{amsmath,url,hyperref}
\usepackage{amsfonts}
\usepackage{amssymb}
\newcommand{\Rr}{{\mathbb R}}

\newcommand{\Cc}{{\mathbb C}}
\newcommand{\Rnt}{R_{\hat{n}_i,\theta_i}}
\newcommand{\ip}[2]{\langle{#1},{#2}\rangle} % Add HASH just before 1 and 2
\begin{document}
\title{Tangled Splines}
\author{Aditya Tatu
\thanks{Dhirubhai Ambani Institute of Information \& Communication Technology, Gandhinagar, India. E-mail: aditya\_tatu@daiict.ac.in.}}% <-this % stops a space
%%\thanks{J. Doe and J. Doe are with Anonymous University.}% <-this % stops a space
%%\thanks{Manuscript received April 19, 2005; revised August 26, 2015.}}
%
%% note the % following the last \IEEEmembership and also \thanks - 
%% these prevent an unwanted space from occurring between the last author name
%% and the end of the author line. i.e., if you had this:
%% 
%% \author{....lastname \thanks{...} \thanks{...} }
%%                     ^------------^------------^----Do not want these spaces!
%%
%% a space would be appended to the last name and could cause every name on that
%% line to be shifted left slightly. This is one of those "LaTeX things". For
%% instance, "\textbf{A} \textbf{B}" will typeset as "A B" not "AB". To get
%% "AB" then you have to do: "\textbf{A}\textbf{B}"
%% \thanks is no different in this regard, so shield the last } of each \thanks
%% that ends a line with a % and do not let a space in before the next \thanks.
%% Spaces after \IEEEmembership other than the last one are OK (and needed) as
%% you are supposed to have spaces between the names. For what it is worth,
%% this is a minor point as most people would not even notice if the said evil
%% space somehow managed to creep in.

\maketitle

\begin{abstract}
Extracting shape information from object boundaries is a well studied problem in vision, and has found tremendous use in applications like object recognition.  Conversely, studying the space of shapes represented by curves satisfying certain constraints is also intriguing. In this paper, we model and analyze the space of  shapes represented by a $3$D curve (space curve) formed by connecting $n$ pieces of quarter of a unit circle. Such a space curve is what we call a \emph{Tangle}, the name coming from a toy built on the same principle. 

We provide two models for the shape space of $n$-link open and closed tangles, and we show that tangles are a subset of trigonometric splines of a certain order. We give algorithms for curve approximation using open/closed tangles, computing geodesics on these shape spaces, and to find the deformation that takes one given tangle to another given tangle, i.e., the Log map. The algorithms provided yield tangles upto a small and acceptable tolerance, as shown by the results given in the paper.
\end{abstract}

%\begin{IEEEkeywords}
\textbf{Keywords:} Tangles, Trigonometric splines, Implicitly defined manifolds, Shape, Space curves, Geodesics, Exponential map, Log map.
%\end{IEEEkeywords}

% \fi
%\IEEEpeerreviewmaketitle
\section{Introduction}
The shape of an object is of tremendous interest to the computer vision community. It has several applications like object recognition, summarizing shape of collection of objects (mean shape), object retrieval and others. There are various aspects and inquiries with respect to shape of an object: shape statistics, shape matching, shape based registration, modeling shape variability and others. 

Shape is typically defined as whatever is left in the object representation after eliminating variability due to rigid transformations\cite{Kendall1989}. Factoring out these variations depends on the mode of object representation chosen. One popular choice is to represent the object by its boundary, which in turn can be represented using one of the following: landmark points, Fourier descriptors, smooth curves, level set functions, medial axis, and others. 

The set of all possible object representation chosen forms a space to work with, from which factoring out the variabilities not of interest gives us a \emph{shape space}. Using tools of differential geometry, one then works with metrics, geodesic distances, parallel transport to answer queries related to shape as listed in the first paragraph. 

Interesting shape (or curve) spaces have been proposed by putting restrictions on variabilities of an object or limiting the object representation, for example, cubic splines, Fourier descriptors, and the space of Bicycle-chain shape models. In this paper, we are interested in studying and analyzing what we call the \emph{Tangle shape space}. Tangle is a toy (and much more) created by Richard X. Zawitz's Tangle creations (\url{www.tanglecreations.com}), consisting of connected copies of a planar curve (arc) to form a closed space curve. Few examples are shown in Figure \ref{fig:1} below. Each arc is rigid and each joint can be subjected to local rotations which may have global effects on the shape. In this paper, we are interested in the mathematical modeling of the tangle shape space, and developing shape related computations on it. We consider two cases, open tangles and closed tangles; in the former the initial point of the first link and final point of the last link need not be the same, while in the latter, the points and the respective tangents must be the same. 

We develop mathematical models for tangles as parametric space curves, and show that the set of all possible tangle curves is a subset of all trigonometric splines of a specific order. Apart from this, some useful characteristics of tangles, curvature and torsion for instance, are also derived. We also show that the set of all possible open tangles can be modeled as an appropriate dimensional torus, while the set of all possible closed tangles is a subset of such a torus. We also provide algorithms for space curve approximation using tangles, and for computing geodesics and Log map on the tangle shape space. Exploring applications like protein shape modeling is left to another paper.
\begin{figure}[ht]
\begin{center}$
\begin{array}{cc}
\includegraphics[width=1.75in]{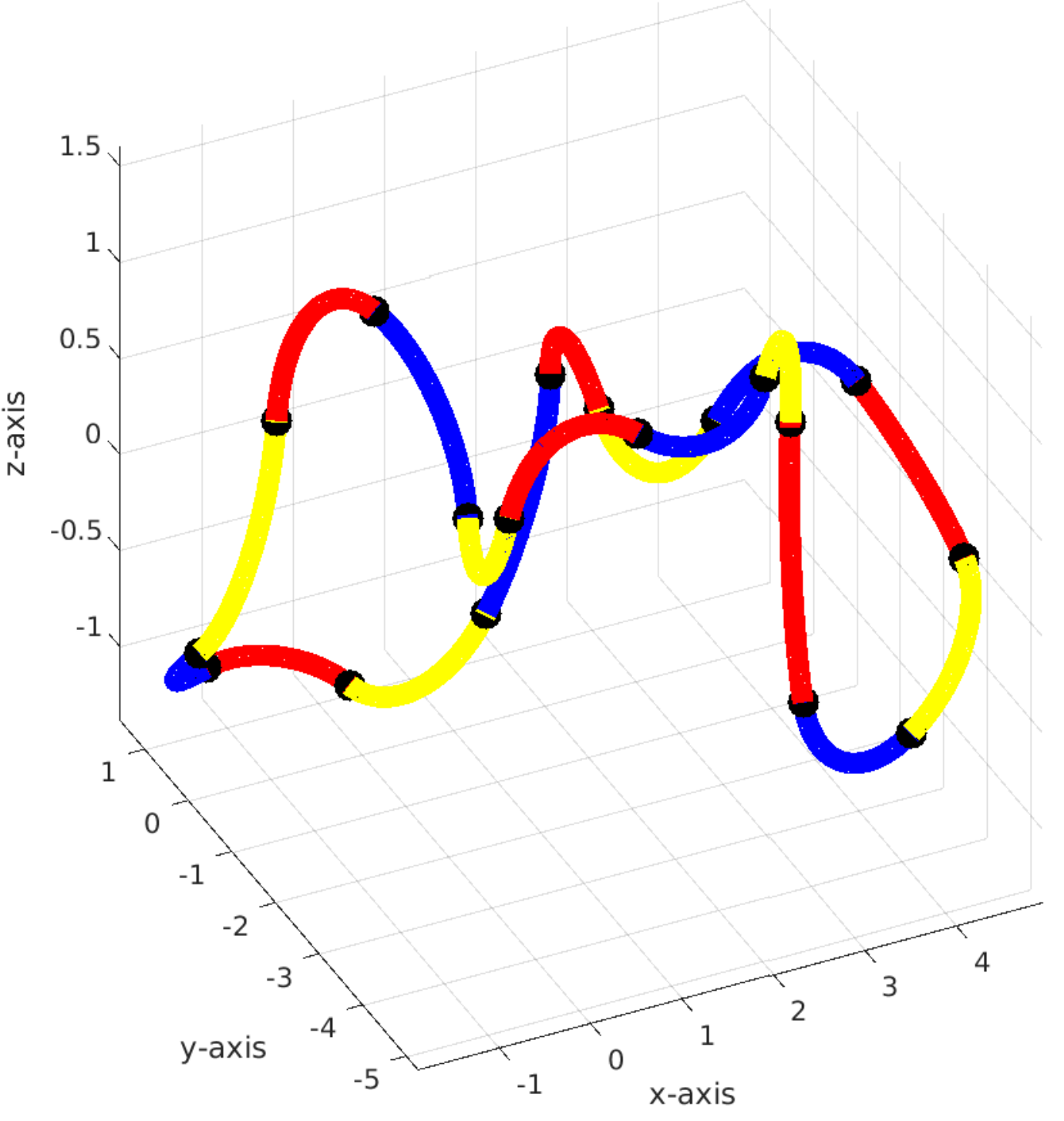} &
\includegraphics[width=1.75in]{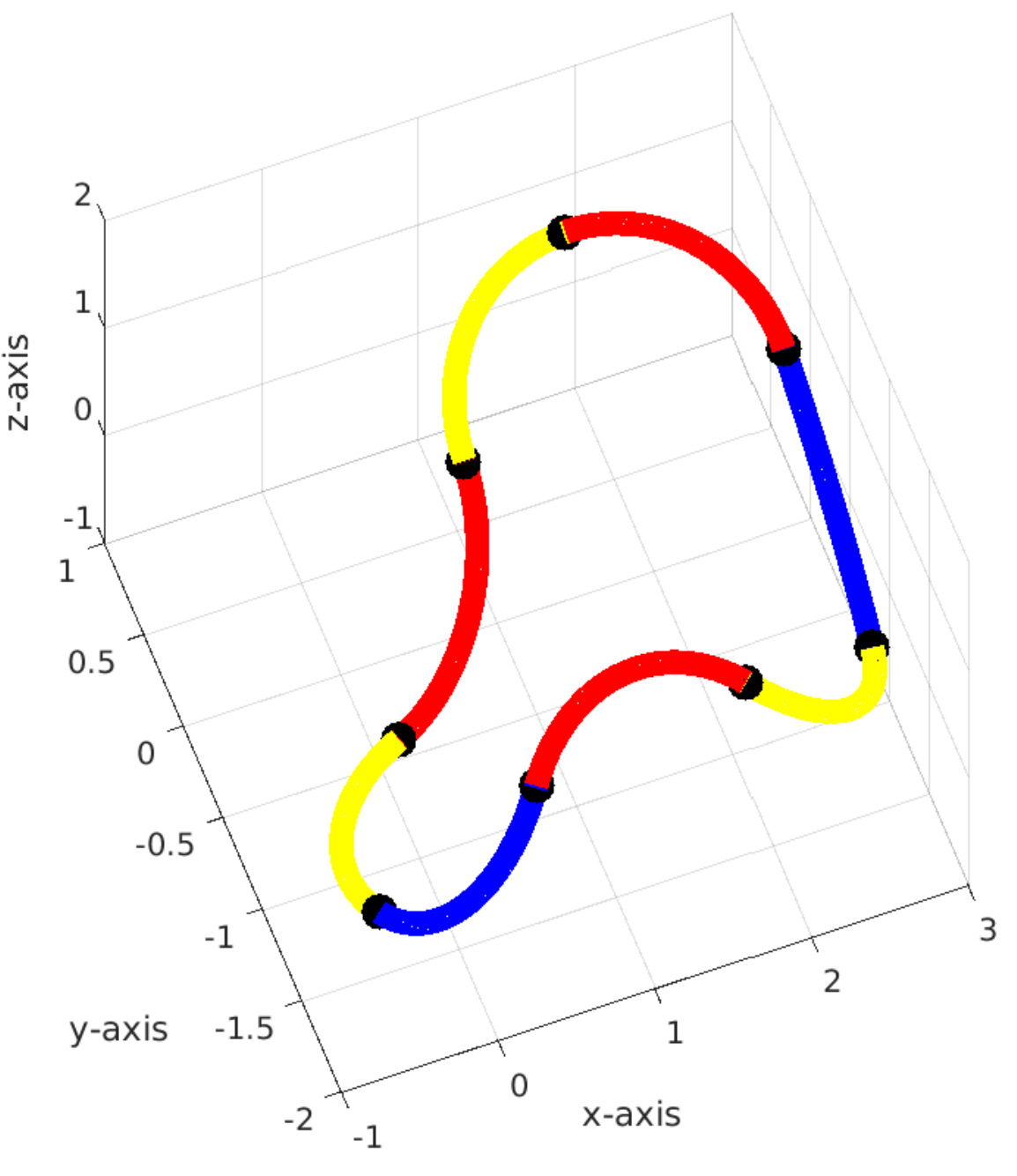}\\
\includegraphics[width=1.75in]{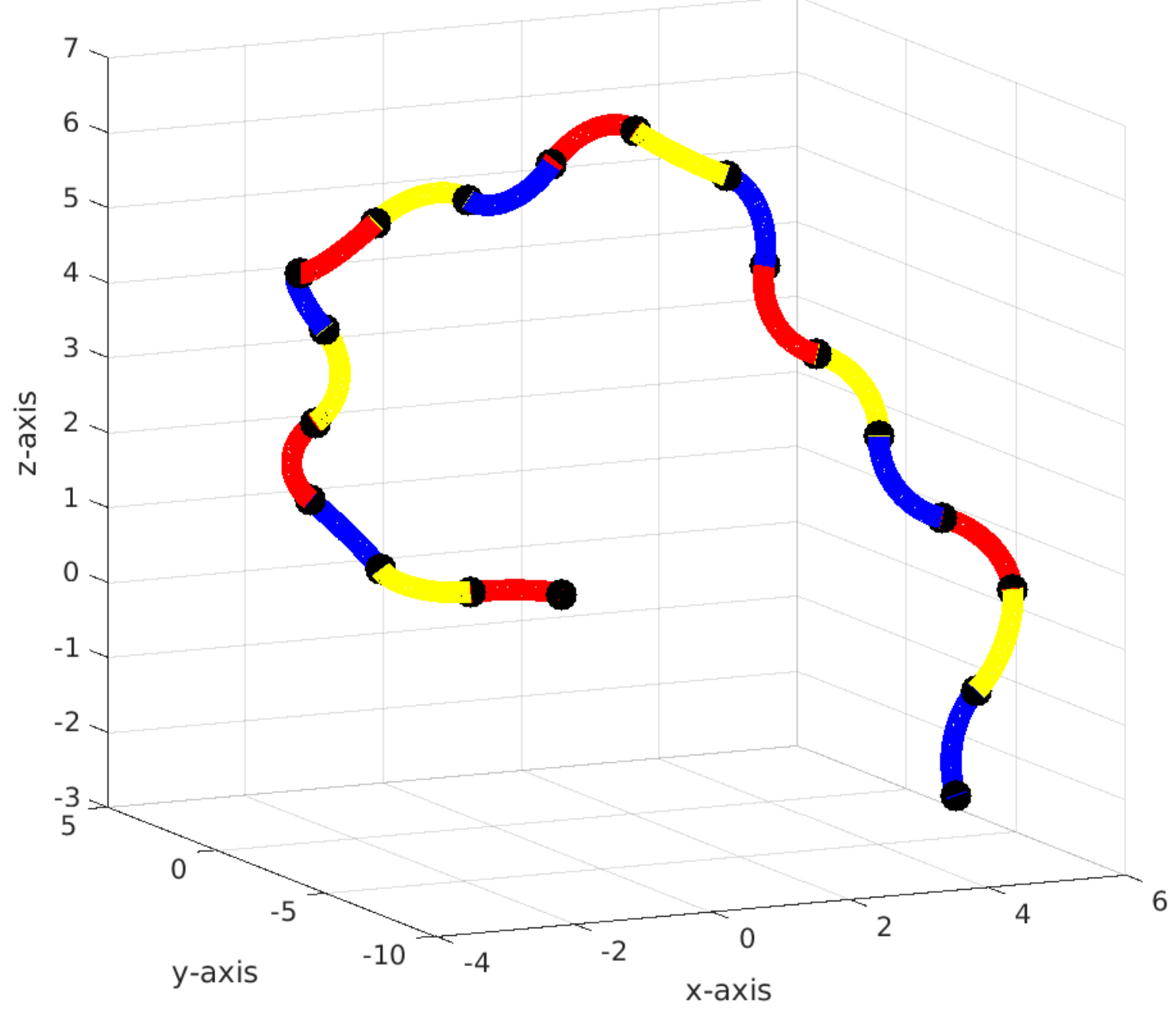} &
\includegraphics[width=1.75in]{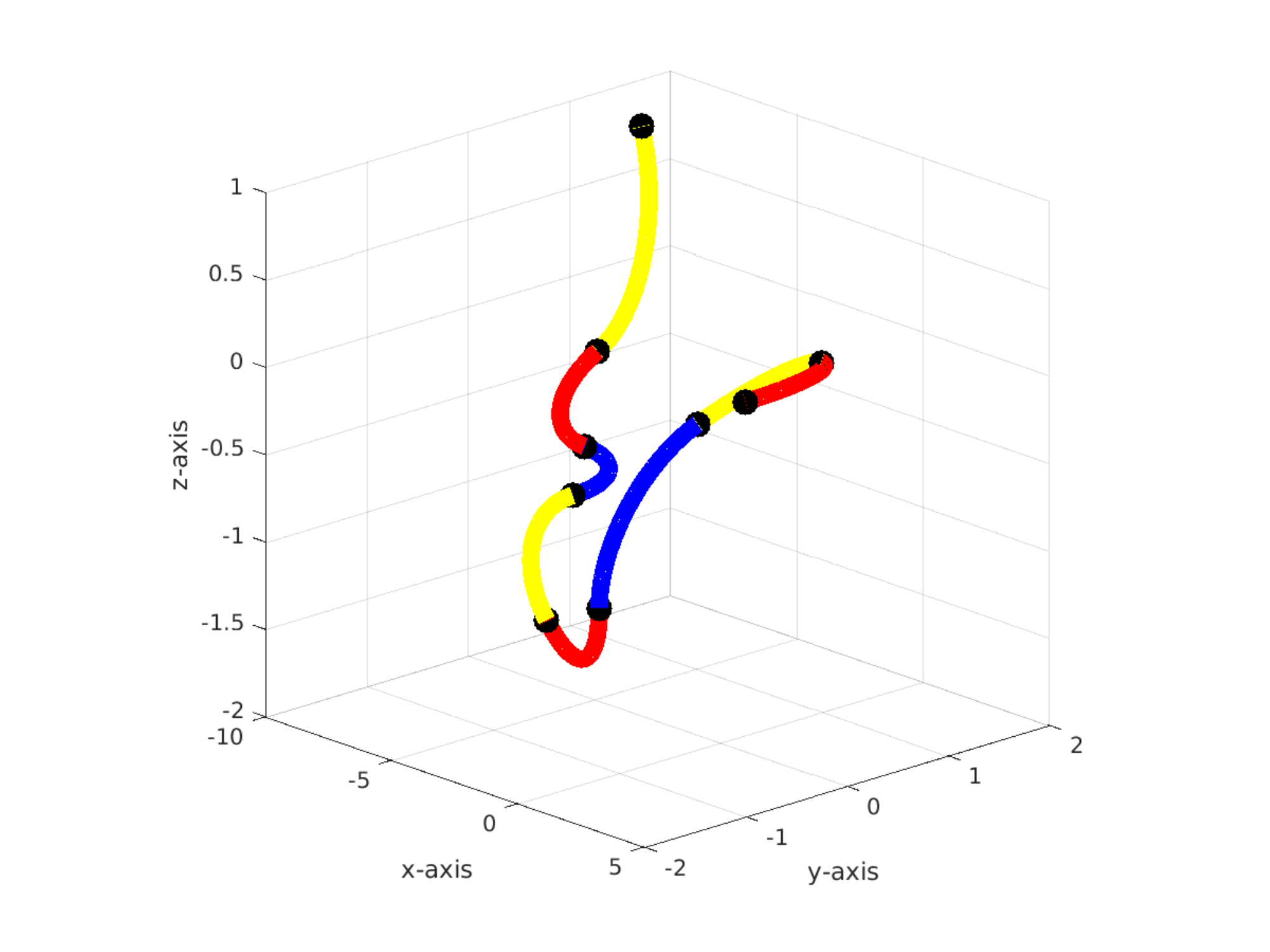}
\end{array}$
\end{center}
\caption{Simulated Tangle examples: (Top) Closed tangles with $18$(left) and $8$(right) links. (Bottom) Open tangles with $18$(left) and $8$(right) links. Note that each distinct color link is a quarter of a unit-circle.}
\label{fig:1}
\end{figure}

The paper is organized as follows: We summarize related work in the next section, and develop models for shape spaces of tangles in Section \ref{sec:modeltangle}. In the same section, we also show that tangles form a subset of trigonometric splines. Thereafter we provide computational algorithms for curve approximation, computing geodesics and Log map on tangle shape space in Section \ref{sec:comptools}. We demonstrate the performance of these algorithms in Section \ref{sec:expts}, and conclude with some discussions in Section \ref{sec:discon}.
\section{Related Work}
The first proper study of shape is credited to D'Arcy Thompson for his work \emph{On Growth and Forms}\cite{Thompson1917}. The modern treatment of shape is mainly due to Kendall\cite{Kendall1984} and Bookstein\cite{Bookstein1986}, wherein it was proposed to represent an object by some landmark points located on the boundary of the object in $\Rr^2$. Kendall used the quotient of this space with the group of translation, scaling and rotation to obtain a \emph{shape space} on which he defined the dissimilarity of two shapes by computing the geodesic distance between the two shapes on the shape space. 

Planar curves have received much more attention than space curves. The space of planar smooth curves is an important model for shape space (though not strictly a shape space). In a more mathematical treatment of this subject, Michor and Mumford\cite{Michor2007} study the set of embeddings and immersions of $S^1$ into $\Rr^2$, where $S^1$ is the unit circle in $\Cc$ and represents the values that a parameter used to define a closed curve can take. The space they consider is then the quotient of the above space with the group of diffeomorphisms from and to $S^1$ representing the group of reparameterizations. This forms an infinite dimensional space on which geodesics are obtained for various metrics. In \cite{Younes1998}, Younnes also works on the space of smooth planar curves, and defines the distance between two curves as the energy needed to deform one curve to another. The energy is based on the length of path on the group of infinitesimal deformation field that deforms one curve to another. 
Klassen et.al.\cite{Klassen2004} use Fourier descriptors of the tangent to the planar curve to represent the curve. A $2$D closed curve can also be represented by its \emph{medial axis}\cite{Blum1973}. A more comprehensive representation called the shock graph has been explored for representation and shape distance computation in \cite{Giblin1999,Sebastian2001}. Sommer et.al. introduced the Bicycle chain shape models\cite{Sommer2009} in which, every landmark on the curve is constrained to be equidistant from its neighbors. Glaunes et\ al.\cite{Glaunes2006} consider a shape as an interior of a planar curve, while the shape space is built by looking at all possible diffeomorphic deformations of the interior of closed curves.

Srivastava et.al. \cite{Srivastava2011} represent curves in $\Rr^n$ using the \emph{square root velocity} function, using which they devise shape analysis tools. Another interesting representation of space curves is presented by Kessler\cite{Kessler2007}, in which a space curve is represented via segments of planar curves called $E$-sets, and has been used for shape analysis of protein molecules. While there exists generalizations of models like the Fourier descriptors to space curves\cite{Reti1993}, splines can be used for space curves without any modification. Splines were introduced by  Schoenberg\cite{Schoenberg1964} and de Boor \cite{Boor1962} as interpolation methods, and have found use in signal \& image processing\cite{Unser1993}, computer graphics\cite{Bartels1987}, manufacturing processes\cite{Weiss1997}, and others. Several types and variations of splines have been developed based on choice of basis and constraints; B-splines, B\'ezier curves, Rational B-splines, Trigonometric splines, to name a few. Apart from applications several theoretical results have also been worked out\cite{Schumaker2007},\cite{Champion2009}. Following \cite{Schumaker2007}, we briefly introduce trigonometric splines here, as the tangle shape space will be shown to be its subset later on.

Given a positive integer $m$, an $m$-dimensional space is defined as 
\[
    \mathcal{T}_m = \left\{
                \begin{array}{ll}
                  span\{v_1,\ldots,v_m\}, \ \ m = 2r\\
				  span\{u_1,\ldots,u_m\}, \ \ m = 2r+1
                \end{array}
              \right.,
\]
where 
\begin{align*}
\{v_1,\ldots,v_m\} = & \{\cos(t/2),\sin(t/2),\ldots\\
                     &\ \cos[(r-1/2)t],\sin[(r-1/2)t]\}, \ \ m = 2r,\\
\{u_1,\ldots,u_m\} = & \{1,\cos(t),\sin(t),\ldots,\\
					 &\ \cos(rt),\sin(rt)\}\ \ m = 2r+1.
\end{align*}
Let $D^k, k \geq 0$ represent the derivative of order $k$ operator. Then, given a partition $\Delta = \{a = t_0 <t_1,\ldots,t_{k+1} = b\}$ of an interval $[a,b]$, and a vector $\mathcal{M} = (m_1,\ldots,m_k)$ of $k$ integers, with $1\leq m_i \leq m$, the space of trigonometric splines of order $m$, with multiplicities $\mathcal{M}$ is defined as 
\begin{align}
  \mathcal{S}(\mathcal{T}_m,\mathcal{M},\Delta) = \left\{
                \begin{array}{ll}
                  s\ |\ \exists \ s_0,\ldots,s_k \in \mathcal{T}_m \mbox{ with } s(t) = s_i(t)\\
                  \forall t \in [t_i,t_{i+1}], i = 0,\ldots,k, \mbox{ and } \\
                  D^{j-1}s_{i-1}(t_i) = D^{j-1}s_i(t_i)\\
                  j = 1,\ldots,m-m_i,\ \ i=1,\ldots,k.
                \end{array}
                \right\}
\label{eqn:trigspl}
\end{align}
In words, a trigonometric spline is a collection of elements from $\mathcal{T}_m$, each defined on a sub-interval $[t_i,t_{i+1}]$, such that at common points of adjacent elements (also called \emph{knots}), they satisfy continuity constraints for derivatives upto order $m-m_i-1$. Thus the space $\mathcal{S}(\mathcal{T}_m,\mathcal{M},\Delta)$ is $m + \sum_{i=1}^k m_i$ dimensional.

Of special interest to us is the space of trigonometric splines of order $m=3$, with $\mathcal{M} = (1,1,\ldots,1)$, henceforth denoted by $\mathcal{M}_1$. We get $\mathcal{T}_3 = \{1,\cos(t),\sin(t)\}$ in this case, while the constraints defined in Equation \eqref{eqn:trigspl} are
\begin{align}
\label{eqn:T3con1}s_{i-1}(t_i) = s_i(t_i),\ \ i=1,\ldots,k\\
\label{eqn:T3con2}Ds_{i-1}(t_i) = Ds_i(t_i),\ \ i=1,\ldots,k.
\end{align}
It can be seen that the space $\mathcal{S}(\mathcal{T}_3,\mathcal{M}_1,\Delta)$ is $k+3$ dimensional. In order to represent a space curve as a trigonometric spline, each coordinate is considered as a trigonometric spline function of the parameter $t$, and the space curve between a sub-interval $[t_i,t_{i+1}]$ of $\Delta$, is given by (for $i=0,\ldots,k$),
\begin{align}
\label{eqn:T3}s_i(t) = a_i + b_i\cos(t-t_i) + c_i\sin(t-t_i),\ \forall t \in [t_i,t_{i+1}],
\end{align}
where $a_i,b_i,c_i \in \Rr^3$ are the coefficient vectors that satisfy constraints given by Equations \eqref{eqn:T3con1},\eqref{eqn:T3con2}. The degrees of freedom available for a trigonometric spline space curve is thus $3k+9$.

In the next section, we model the Tangle shape space and show that it is a subset of $\mathcal{S}(\mathcal{T}_3,\mathcal{M}_1,\Delta)$.
%\IEEEpubidadjcol
\section{Modeling Tangles}
\label{sec:modeltangle} In this work, we will represent tangle configurations using parametric space curves, formed by connecting $n$ copies of a quarter of a unit-circle (each copy called a link), translated and rotated appropriately. Depending on whether the first and last links are connected or not, we obtain a closed or an open tangle. Different configurations are obtained with the help of local rotations applied on each link. These local rotations have a global effect on the tangle, akin to the effect of changing a control point of a spline curve.
In what follows, $s$ represents a parametric space curve, $s(t)$ denotes the point on the curve at parameter value $t$, while $\dot{s}(t)$ denotes the derivative of the functions $s$ at parameter value $t$, and represents the tangent to the curve at point $s(t)$. Unless stated otherwise, $(p,q,r) \in \Rr^3$ represents a row vector, while $(p,q,r)^T \in \Rr^3$ or any element $v \in \Rr^3$ will be assumed to be represented as column vectors.

Since each link is a translated and rotated version of a quarter  of a unit-circle, without loss of generality, we assume the following canonical form of a quarter of a unit-circle which when translated and rotated appropriately, yields any given link:
\begin{equation}
\label{eqn:can}c(t) = (\cos t,\sin t,0)^T,\ \forall t \in [0,\frac{\pi}{2}].
\end{equation}  
The tangent to $c(t)$ is $\dot{c}(t) = (-\sin t,\cos t,0)^T,\ \forall t \in [0,\frac{\pi}{2}]$, specifically, $\dot{c}(0) = (0,1,0)^T$ and $\dot{c}(\frac{\pi}{2}) = (-1,0,0)^T$. Let $s(t), t \in [0,n\frac{\pi}{2}]$ denote the $n$-link space curve, and let $t_i = \frac{i\pi}{2}, i = 0,\ldots,n$.
Let $R_{\hat{n},\theta} \in \Rr^{3 \times 3}$ denote the rotation matrix with axis $\hat{n}$ and angle of rotation $\theta$. For a given space curve $s$ that represents an $n$-link tangle, let $s_i(t) = s(t),t \in [t_i,t_{i+1}]$ denote the $(i+1)^{st}$ link, where $i=0,\ldots,n-1$. Since any link is a rotated and translated form of the assumed canonical form $c$ from Equation \eqref{eqn:can}, we have,
\begin{align}
\label{eqn:sit}s_i(t) =& \Rnt c(t-t_i) + T_i,\ \forall t \in [t_i,t_{i+1}],
\end{align}
for some unit normal $\hat{n}_i$, angle $\theta_i$, and translation vector $T_i \in \Rr^3$. The tangent to the $(i+1)^{st}$ link is 
\begin{align}
\dot{s}_i(t) =& \Rnt \dot{c}(t-t_i), \forall t \in [t_i,t_{i+1}],
\end{align}
and specifically, 
\begin{align}
\label{eqn:1}\dot{s}_i(t_i) =& \Rnt (0,1,0)^T\\
\label{eqn:2}\dot{s}_i(t_{i+1}) =& \Rnt (-1,0,0)^T.
\end{align}
Moreover, since $\ip{\dot{c}(0)}{\dot{c}(\frac{\pi}{2})} = 0$, $\ip{\dot{s}_i(t_i)}{\dot{s}_i(t_{i+1})} = 0$, and $||\dot{c}(0)|| = ||\dot{c}(\frac{\pi}{2})|| = ||\dot{s}_i(t_i)|| = ||\dot{s}_i(t_{i+1})|| = 1$.

Thus, one can determine $\Rnt$ uniquely from an orthogonal pair of unit vectors $(\dot{s}_i(t_i), \dot{s}_i(t_{i+1}))$ by solving Equations \eqref{eqn:1},\eqref{eqn:2}, and conversely, given $\Rnt$, one can determine the tangent vectors: $\dot{s}_i(t_i), \dot{s}_i(t_{i+1})$ from the same equations. Given two consecutive links $s_i$ and $s_{i+1}$, these should satisfy the following constraints:
\begin{align}
\label{eqn:tancon1}s_i(t_{i+1}) =& s_{i+1}(t_{i+1})\\
\label{eqn:tancon2}\dot{s}_i(t_{i+1}) =& \dot{s}_{i+1}(t_{i+1}),
\end{align} 
the former stating that the end point of link $(i+1)$, $s_i(t_{i+1})$ should be the same as the starting point of link $i+2$, $s_{i+1}(t_{i+1})$, and the latter ensuring that the tangent vector at the end point of link $i+1$, $\dot{s}_i(t_{i+1})$ should be the same as the tangent vector at the starting point of link $i+2$, $\dot{s}_{i+1}(t_{i+1})$. The first constraint gives:
\begin{align*}
\Rnt c(\frac{\pi}{2}) + T_i = R_{\hat{n}_{i+1},\theta_{i+1}} c(0) + T_{i+1}.
\end{align*}
Thus, if $s_i(t)$ is known and $R_{\hat{n}_{i+1},\theta_{i+1}}$ is known (using tangent vectors as discussed earlier), $T_{i+1}$ can be computed using
\begin{align}
\label{eqn:ti}T_{i+1} = \Rnt c(\frac{\pi}{2}) + T_i - R_{\hat{n}_{i+1},\theta_{i+1}} c(0).
\end{align}
Let $V_i, V_{i+1}, V_{i+2}$ denote the tangent vectors $\dot{s}_i(t_i), \dot{s}_i(t_{i+1}) = \dot{s}_{i+1}(t_{i+1}), \dot{s}_{i+1}(t_{i+2})$ respectively. Since $\ip{V_i}{V_{i+1}} = \ip{V_{i+1}}{V_{i+2}} = 0$, one can represent the tangent vector $V_{i+2}$ as $V_{i+2} = R_{V_{i+1},\theta_i} V_i$, for some $\theta_i$. In words, given the orthonormal tangent vectors $V_i, V_{i+1}$, any unit vector $V_{i+2}$ which is orthogonal to $V_{i+1}$ can be produced via a rotation about the axis $V_{i+1}$ of the vector $V_i$.

Let $V_i = \dot{s}(t_i)$ denote the tangent vector at the initial point of $(i+1)^{st}$ link for $i = 0,\ldots,n-1$, and $V_n = \dot{s}(t_n)$ denote the tangent vector at the end-point of the $n^{th}$ link. Assuming $V_0,V_1$ to be given such that $\ip{V_0}{V_1} = 0, ||V_0|| = ||V_1|| = 1$, an $n$-link \emph{open} tangle parametric space curve $s:[0,n\frac{\pi}{2}] \rightarrow \Rr^3$, is also uniquely represented by the set of angles $(\theta_0,\ldots,\theta_{n-2})$, where $V_{i+2} = R_{V_{i+1},\theta_i} V_i, i = 0,\ldots,n-2$. 
Thus, including the degrees of freedom in choosing the orthonormal vectors $V_0,V_1$ (which is three), and degrees of freedom available through a global translation (again three), the total degrees of freedom for an $n$-link open tangle is $n+5$.

A closed form expression for each $s_i(t)$ can be obtained by carefully observing Equations \eqref{eqn:1},\eqref{eqn:2}. It is not difficult to see that the first two columns of $\Rnt$ in Equation \eqref{eqn:sit} are $-s_i(t_{i+1})$ or $-V_{i+1}$ and $s_i(t_i)$ or $V_i$. Substituting these in Equation \eqref{eqn:sit} and \eqref{eqn:ti}, and noting the special form of $c$ from Equation \eqref{eqn:can}, one gets the following closed form expression:
\begin{align}
\label{eqn:clsit}s_i(t) = & V_i\sin(t-t_i) - V_{i+1}\cos(t-t_i) + T_i, \forall t \in [t_i,t_{i+1}],
\end{align}
for $i = 0,\ldots,n-1$, and 
\begin{align}
T_{i+1} =& V_i + V_{i+2} + T_i, i = 0,\ldots,n-2,
\end{align}
with $T_0 = 0$. Comparing Equations \eqref{eqn:T3con1},\eqref{eqn:T3con2} and \eqref{eqn:T3} with Equations \eqref{eqn:tancon1},\eqref{eqn:tancon2} and \eqref{eqn:clsit}, it is clear that the set of all tangle curves is a subset of $\mathcal{S}(\mathcal{T}_3,\mathcal{M}_1,\Delta)$.

In case shape as defined by Kendall\cite{Kendall1989} of an $n$-link open tangle is concerned, the degrees of freedom reduces (by six) to $n-1$. As far as shape representation for an $n$-link open tangle is of interest, one can factor out the global translation and rotation by fixing the first of the $n$ links to our canonical quarter circle $c$ from Equation \eqref{eqn:can}, i.e., $s_0(t) = c(t), t \in [0,\frac{\pi}{2}]$. This implies fixing $V_0 = (0,1,0)^T$ and $V_1 = (-1,0,0)^T$. The angles $(\theta_0,\ldots,\theta_{n-2})$ are the degrees of freedom that generate $n$-link open tangles with different shapes\footnote{Some $n$-link tangles may have additional symmetries which we ignore in this work.}, and thus the shape space of an $n$-link open tangle is the $n-1$ dimensional torus $\mathbb{T}^{n-1}$.

Thus the set of $n$-link open tangle shapes $S_o$, can be modeled by any of the following equivalent ways:
\begin{equation}
\begin{gathered}
S_o = \mathbb{T}^{n-1}\\
\mbox{ or }\\
S_o = \left\{(V_2,\ldots,V_n), V_i \in \Rr^3, i = 2,\ldots,n\ |\ \ip{V_i}{V_{i+1}} = 0,\right.\\
\left. i = 2,\ldots,n-1, ||V_i|| = 1, i=2,\ldots,n\right\}
\end{gathered}
\end{equation}
with $V_0 = (0,1,0)^T$ and $V_1 = (-1,0,0)^T$.
%\begin{eqnarray*}
%\begin{gather*} 
%\mathbb{T}^{n-1}\\
%\mbox{ or }\\
%&\left\{(V_2,\ldots,V_n), V_i \in \Rr^3, i = 2,\ldots,n\ |\ \ip{V_i}{V_{i+1}} = 0,\right.\\
%&\left. i = 2,\ldots,n-1, \ip{V_i}{V_i} = 1, i=2,\ldots,n\right\}
%\end{gather*}
%\end{eqnarray*}
\subsection{$n$-link closed tangles}
For $s:[0,n\frac{\pi}{2}]\rightarrow \Rr^3$ to represent a closed $n$-link tangle, it needs to satisfy two additional constraints:
\begin{align}
s(t_n) =& s(t_0)
V_n =& V_0\\
\end{align}
The first constraint ensures that the first and the last point coincide, while the second ensures continuity of the first derivative. The first condition can also be met by using the fact that line integral over a closed curve of its tangent vector field is zero, which in terms of $s$ can be written and simplified as,
\begin{align}
\int_0^{n\frac{\pi}{2}} \dot{s}(t)\ dt & = \sum_{i=0}^{n-1} \int_{i\frac{\pi}{2}}^{(i+1)\frac{\pi}{2}} \dot{s}_i(t)\ dt\\
%									  & = \sum_{i=0}^{n-1} s_i((i+1)\frac{\pi}{2}) - s_i(i\frac{\pi}{2})\\
									  & = \sum_{i=0}^{n-1} \Rnt \left(c(\frac{\pi}{2}) - c(0)\right)\\
									  & = \sum_{i=0}^{n-1} \Rnt \left(\dot{c}(\frac{\pi}{2}) + \dot{c}(0)\right) = 0,								  
\end{align}
the last equality resulting because of the special choice of $c$.
Using Equations \eqref{eqn:1},\eqref{eqn:2}, and representing the tangent vectors with $V_i$, the constraint that the initial and end-point of the $n$-link tangle must coincide boils down to,
\begin{align}
\sum_{i=0}^{n-1} V_i = 0.
\end{align}
In case shape of the $n$-link closed tangle is of interest, we fix $V_0 = (0,1,0)^T$ and $V_1 = (-1,0,0)^T$. We thus need to ensure $\ip{V_0}{V_{n-1}} = 0$. With $V_{i+2} = R_{V_{i+1},\theta_i}, i = 0,\ldots,n-3$, the $n$-link closed tangle shape space is $S_c = F^{-1}(0)$, where $F:\mathbb{T}^{n-2} \rightarrow \Rr^4$ is defined as, 
\begin{align}
F(\theta_0,\ldots,\theta_{n-3}) = \left(\sum_{i=0}^{n-1} V_i,\ip{V_0}{V_{n-1}}\right).
\end{align} 
Thus, the degrees of freedom\footnote{The reason behind not using the term dimension in place of degree of freedom will be made clear in a later section.} as far as shapes of $n$-link closed tangles is $n-6$, and the shape space is a subset of $\mathbb{T}^{n-2}$. The $n$-link closed tangle shapes $S_c$, can be modeled by any of the following equivalent ways:
\begin{equation}
\begin{gathered}
S_c = \left\{(\theta_0,\ldots,\theta_{n-3}) \in \mathbb{T}^{n-2}\ \left|\right.\ F(\theta_0,\ldots,\theta_{n-3}) = 0\right\}\\
\mbox{ or }\\
S_c = \left\{(V_2,\ldots,V_{n-1}), V_i \in \Rr^3, i = 2,\ldots,n-1\ |\right.\\ 
\ip{V_i}{V_{i+1}} = \ip{V_0}{V_{n-1}} = 0, i = 1,\ldots,n-2,\\
\left.||V_i|| = 1, i=2,\ldots,n-1, \sum_{i=0}^{n-1} V_i = 0\right\},
\end{gathered}
\end{equation}
with $V_0 = (0,1,0)^T$ and $V_1 = (-1,0,0)^T$.
For computations on the tangle shape space, we will prefer the latter characterizations in terms of unit tangent vector sets (satisfying appropriate constraints), instead of elements of the appropriate torus, since closed form expressions relating the angle $\theta_i$ with the unit tangent vector $V_i$ become far too involved as the number of links increases. This is especially true for closed tangles.

As expected, the set of $4$-link closed tangle shape space contains one point, the circle starting with the canonical quarter $c$. It can also be shown that there are no closed tangle curves with $n=5$ links\footnote{We avoid giving the proof here as it lacks elegance and it takes the paper in a tangential direction}.
\subsection{Properties of Tangles}
We now briefly summarize the properties of tangles as space curves.
\begin{enumerate}
\item \emph{Unit Speed curves:} Tangles as given by Equations \eqref{eqn:clsit}, \eqref{eqn:can} are unit-speed parametric space curves. This is obvious, since for the canonical form $||\dot{c}(t)|| = 1, \forall t$, while all links are obtained by rotating and translating this canonical form.
\item \emph{Curvature:} The tangent at any point $t \in [t_i,t_{i+1}]$ is given by
\begin{align}
\dot{s}_i(t) = V_i\cos(t-t_i) + V_{i+1}\sin(t-t_i).
\end{align}
Note that the space of tangle curves was shown to be a subset of $\mathcal{S}(\mathcal{T}_3,\mathcal{M}_1,\Delta)$, hence second order derivatives may not exist at the knots/points connecting two links. But at all other points, the tangles are smooth. Since the curve is unit speed, the curvature magnitude is $|\kappa(t)| = ||\ddot{s}_i(t)||, \forall t \neq t_i$,
\begin{align}
|\kappa(t)| = ||V_{i+1}\cos(t-t_i) - V_i\sin(t-t_i)|| = 1, \forall t \neq t_i
\end{align}
and therefore $\kappa(t) = \pm 1, \forall t \neq t_i$. 
\item \emph{Torsion:} The unit normal at $t \in [t_i,t_{i+1}]$, is
\begin{align}
\mathbf{n}(t) = V_{i+1}\cos(t-t_i) - V_i\sin(t-t_i).
\end{align}
and therefore the \emph{binormal vector} at $t \in [t_i,t_{i+1}]$, is 
\begin{align*}
\mathbf{b}(t) = V_i \times V_{i+1},
\end{align*}
Since each link lies in the plane spanned by $V_i$ and $V_{i+1}$, for each $t \in (t_i,t_{i+1})$, the torsion $\tau$ is zero.
\end{enumerate}
\section{Computational Tools}
\label{sec:comptools}
We first provide an algorithm to approximate a given space curve with open tangles, followed by algorithms to compute geodesics and Log map in the open/closed tangle shape space.
\subsection{Curve approximation using Tangles}
We assume that we are given a smooth (at least $\mathcal{C}^1$) constant speed parameterized space curve $p:\left[0,n\frac{\pi}{2}\right] \rightarrow \Rr^3$ which is to be approximated by an $n$-link open tangle $s$. In case only an ordered set of points in $\Rr^3$ is to be fit with a tangle, we fit a spline $p$ through the set of points. Our method tries to fit the tangent field of the tangle to the tangent field of the given curve in the least square sense, i.e., it minimizes the following cost:
\begin{align}
\tilde{J}(s) = \int_0^{n\frac{\pi}{2}} ||\dot{p}(t) - \dot{s}(t)||^2\ dt
\end{align}
The tangent field of the given space curve is either computed analytically if the parametric form is available, or computationally via a spline fit as discussed earlier.
This gives,
\begin{align}
\tilde{J}(s) = \sum_{i=0}^{n-1} \int_{t_i}^{t_{i+1}} ||\dot{p}(t) - V_i\cos(t-t_i) - V_{i+1}\sin(t-t_i)||^2\ dt.
\end{align}
Since the tangle is uniquely specified (upto global rigid motions) once the tangent vectors at the connecting points are determined, we obtain a constrained minimization problem in terms of $V = [V_0,\ldots,V_n]$ (note here we do not fix $V_0$ and $V_1$):
\begin{align}
J(V) = & \sum_{i=0}^{n-1} \int_{t_i}^{t_{i+1}} ||\dot{p}(t) - V_i\cos(t-t_i) - V_{i+1}\sin(t-t_i)||^2\ dt \\
\mbox{such that } & \ip{V_i}{V_{i+1}} = 0, i=0,\ldots,n-1,\\
\mbox{and } & ||V_i|| = 1, i = 0,\ldots,n,
\end{align}
Sampling the tangent field of the given space curve $p$ and the tangle curve $s$ to be estimated uniformly (using methods from \cite{Wang2002}, if required), and replacing the integral with a summation at these sample points in the above cost function $J$, gives us a constrained least square problem in the variables $V$, for which we deploy the penalty method from \cite{Luenberger2015}\cite{Bertsekas2009}. Note that since only the tangent fields are aligned, an additional rigid (and global) transformation alignment is performed once the above optimization process ends. 
%In case scale of a configuration is quotiented out, the tangle stops being a unit-speed curve and each link becomes a quarter circle with a radius equal to the scale normalization. 

Given a closed parametric space curve, we can approximate it with a closed tangle configuration with the same method given above, with few additional constraints listed in Section \ref{sec:modeltangle} ensuring that the tangle configuration is closed. 
\subsection{Geodesics on Tangle space}
We now compute geodesics emanating at a given point in the tangle shape space in any given tangent direction. Geodesics on a Riemannian manifold $M$, can be computed via the \emph{Riemannian Exponential map} (henceforth Exp map) as follows. For $p \in M$, and tangent space of $M$ at $p$ denoted by $T_pM$, the Exp map $Exp_p:T_pM \rightarrow M$, maps every $v \in T_pM$ to a point $q \in M$ obtained by moving along a geodesic starting at $p$ in the direction $\frac{v}{||v||}$, for a distance $||v||$, where $||.||$ is the Riemannian metric defined on $T_pM$. Thus given the Exp map $Exp$, one can compute segment of a geodesic $\gamma$ starting at $p$ along a tangent vector $v \in T_pM$ as $\gamma: t \mapsto Exp_p(tv)$, $t \in [0,1]$.

As discussed earlier, the $n$-link open tangle shape space $S_o$ is simply $\mathbb{T}^{n-1}$. Moreover, $\mathbb{T}^{n-1}$ is a Lie group\cite{Boothby2003}, with $T_p(\mathbb{T}^{n-1}) = \Rr^{n-1}$ for any $p \in \mathbb{T}^{n-1}$. Identifying $\mathbb{T}^{n-1}$ with $S^{n-1}$, where $S$ is the unit circle in $\Cc$, we can use the Exp map defined on $S^{n-1}$ to compute geodesics on $\mathbb{T}^{n-1}$. For every $p = (p_0,\ldots,p_{n-2}) \in S^{n-1}$ and $\forall \theta = (\theta_0,\ldots,\theta_{n-2}) \in \Rr^{n-1}$, $Exp_p:\Rr^{n-1} \rightarrow S^{n-1}$ is defined as,
\begin{align}
Exp_p(\theta) = (p_0\exp(i\theta_0),\ldots,p_{n-2}\exp(i\theta_{n-2})),
\label{eqn:expso}
\end{align}
where $i = \sqrt{-1}$, and $\exp$ is the usual exponential function.

The Exp map on the space of closed tangles is more involved. We defer the discussion on whether the set of $n$-link closed tangle configurations forms a manifold or not to a later section. Here, we work under the assumption that geodesics are computed at points contained in a sufficiently large neighborhood where singularities do not occur. The geodesics can be computed in two ways, using each of the model presented in Section\ref{sec:modeltangle}. For brevity, we present the method using the following model for $n$-link closed tangle shapes:
\begin{align}
\nonumber S_c = \left\{(V_2,\ldots,V_{n-1}), V_i \in \Rr^3, i = 2,\ldots,n-1\ |\right.\\ 
\nonumber \ip{V_i}{V_{i+1}} = \ip{V_0}{V_{n-1}} = 0, i = 1,\ldots,n-2,\\
\label{eqn:Sc}\left.||V_i|| = 1, i=2,\ldots,n-1, \sum_{i=0}^{n-1} V_i = 0\right\},
\end{align}
with $V_0 = (0,1,0)^T, V_1 = (-1,0,0)^T$ fixing the first link. Then $S_c = F^{-1}(0)$ is assumed to be an implicitly defined manifold, using the function $F$, where $F:\Rr^{3n-6} \rightarrow \Rr^{2n}$ encode the $2n$ constraints on the set of tangent vectors $v = (V_2,\ldots,V_{n-1}) \in \Rr^{3n-6}$ given in Equation\eqref{eqn:Sc}. To simplify notations, let $p = 3n-6$, and $q = 2n$. The tangent space of $S_c$ at $v$ is given by $ker(JF(v))$, where $ker(A)$ denotes the kernel of matrix $A$, while $JF(v) \in \Rr^{q \times p}$ is the Jacobian matrix of $F$. Let $HF_i(v) \in \Rr^{p \times p}$ denote the Hessian of the constraint $F_i, i = 1,\ldots,q$. Following an optimal control strategy given by Dedieu \& Nowicki \cite{Dedieu2005}, in order to compute the exponential map at a point $v \in S_c$ of the tangent vector $u \in T_v(S_c)$, we solve the following set of differential equations in $v(t), t \in [0,1]$,:
\begin{align}
\nonumber \dot{r}(t) = -\sum_{i=1}^q \mu_i HF_i(v(t)) \dot{v}(t),\\
\nonumber \dot{v}(t) = \left(Id - JF(v(t))^{\dagger} JF(v(t))\right) r(t),\\
\nonumber \mu(t) = - \left(JF(v(t))^T\right)^{\dagger} r(t),\\
v(0) = v, r(0) = u,
\end{align}
where $r \in \Rr^p$ acts as an auxiliary variable, $A^{\dagger}$ is the pseudo-inverse of matrix $A$, and $A^T$ denotes transpose of $A$. This differential equation in the variables $(r,v)$ can be solved using explicit methods or implicit methods given in\cite{Dedieu2005}. We use explicit methods, since we are interested in solving the differential equations for a small interval $t \in [0,1]$, for which in practice the errors are negligible.
\subsection{Log map on Tangle shape space}
We now compute the inverse of the Exp map, the Riemannian Log map. The Exp map is invertible only on a small open set of the manifold known as the cut locus. 

For the $n$-link open tangle space $S_o$, since the Riemannian Exp map is given as Equation \eqref{eqn:expso}, it can be seen that the Log map at $p \in S^{n-1}$ for point $q \in S^{n-1}$, if it exists, is given as, 
\begin{align}
(\theta_0,\ldots,\theta_{n-2}) = \left(\arg\left(\frac{q_0}{p_0}\right),\ldots,\arg\left(\frac{q_{n-2}}{p_{n-2}}\right)\right) \in T_p(S_o),
\label{eqn:logso}
\end{align}
For the $n$-link closed tangle shape space $S_c$, we use the shooting method \cite{Klassen2004},\cite{Mio2004} to compute the Riemannian Log map. Given $v_0 \in S_c$, in order to compute $a = Log_{v_0}(v_1) \in T_{v_0}(S_c)$ for a point $v_1 \in S_c$, the iterative method starts with an initial estimate $\hat{a}_0 = \Pi_{v_0}(v_1 - v_0)$, where $\Pi_{v_0}$ is the orthogonal projection operator to $T_{v_0}(S_c)$. Let $w_k = Exp_{v_0}(\hat{a}_k)$ be the point obtained by \emph{shooting} with the current estimate $\hat{a}_k$. The estimate is updated using the parallel transport to $T_{v_0}(S_c)$ of the projection of the current error, i.e, $\hat{a}_{k+1} = \hat{a}_k + P\left(\Pi_{w_k}(v_1 - w_k)\right)$, where $P$ denotes the parallel transport to $T_{v_0}$ along the geodesic $\gamma(-t), t \in [0,1]$, where $\gamma(t)$ is the geodesic connecting $v_0$ and $w_k$ computed via the Exp map. This is repeated until convergence. 

Let us briefly describe the parallel transport mechanism for transporting $r \in T_{w}(S_c)$ along a geodesic $\gamma \in S_c$ with $\gamma(0) = w, \gamma(1) = v_0$, to a vector $P(r) \in T_{v_0}(S_c)$. Let $r(0) = r$ and $r(1) = P(r)$. The conditions defining such a parallel transport are:
\begin{align*}
& r(t) \in T_{\gamma(t)}(S_c),\ t \in [0,1],\\
& \Pi_{\gamma(t)}\left(\frac{dr}{dt}(t)\right) = 0 \Rightarrow \frac{dr}{dt}(t) \in \left(T_{\gamma(t)}(S_c)\right)^{\perp},\ t \in [0,1].
\end{align*}
While the first equation above states that all vectors $r(t)$ should belong to the appropriate tangent space, the second states that the vector field $r$ obtained via parallel transport should be the one with a zero tangential acceleration along $\gamma$. Rewriting these two equations, we get for all $t \in [0,1]$,
\begin{align*}
JF(\gamma(t))r(t) = 0,\\
\frac{dr}{dt}(t) = JF(\gamma(t))^T\lambda(t),
\end{align*}
where $\lambda \in \Rr^q$. Differentiating the first equation above with respect to $t$, using the second equation above and some algebra gives us the following differential equation in $r$:
\begin{align}
\label{eqn:pt} \frac{dr}{dt}(t) = -JF^T\cdot (JF^T)^{\dagger}\cdot JF^{\dagger} \cdot D^2F(\dot{\gamma},r),
\end{align}
where dependence on $t$ of the right hand side has been suppressed to simplify the expression, and $D^2F(\dot{\gamma},r)$ is a vector in $\Rr^q$ whose $k^{th}$ component ($1\leq k \leq q$) is given by $\dot{\gamma}(t)^T HF_k(\gamma(t)) r(t)$. The above differential equation is solved using an explicit solver to compute $P(r) = r(1)$, for the given $r(0) = r$ and geodesic $\gamma$. 

Apart from the shooting method, one may also use the path straightening approach\cite{Noakes1998} to compute the Log map.
\section{Experiments}
\label{sec:expts}
We now demonstrate results for the computational tools developed in the last section. The algorithms described in the previous section have been implemented in MATLAB.
\subsection{Curve approximation}
\begin{figure*}[t]
 \begin{center}$
 \begin{array}{cc}
 \includegraphics[width=2in]{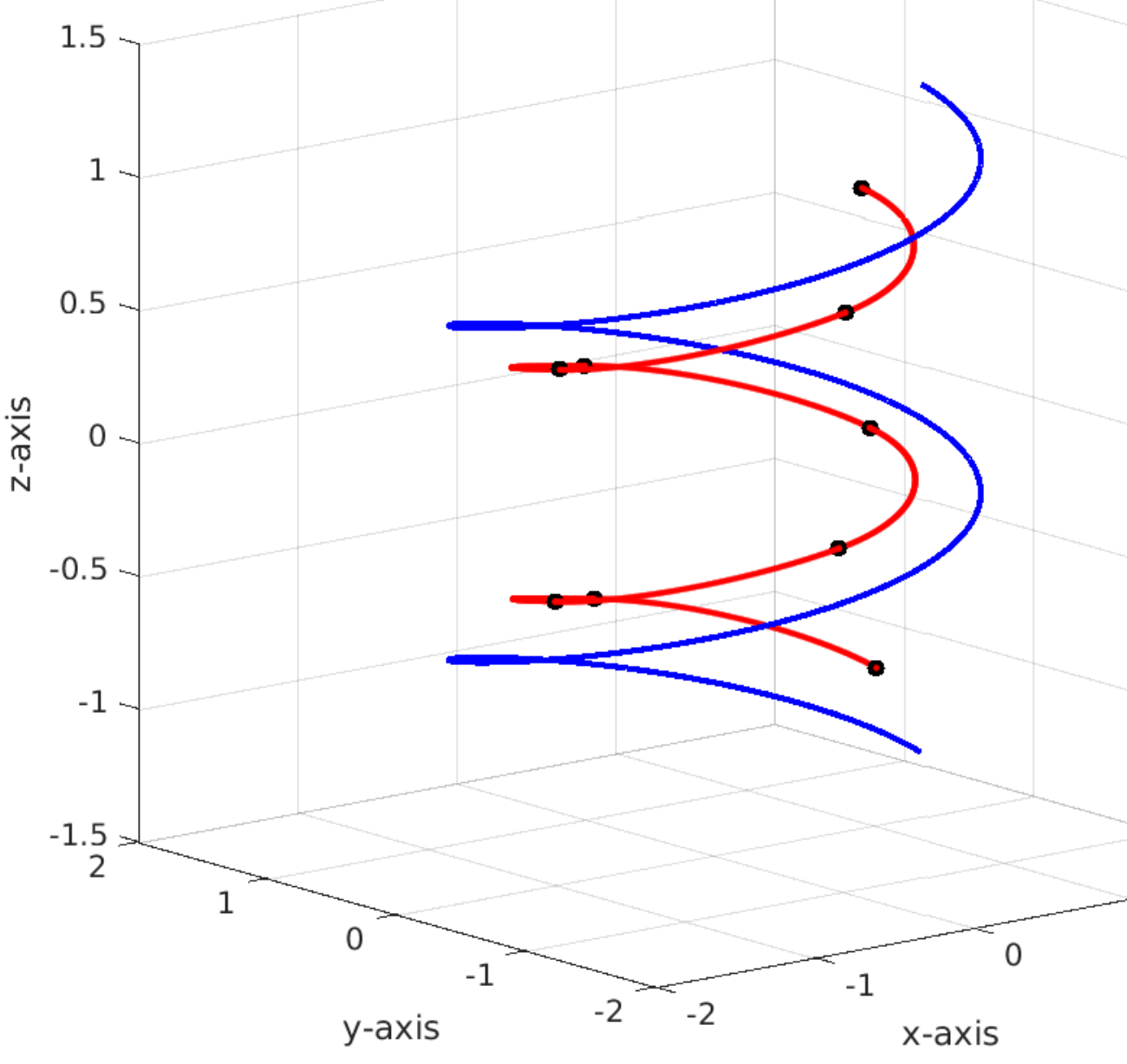} &
 \includegraphics[width=2in]{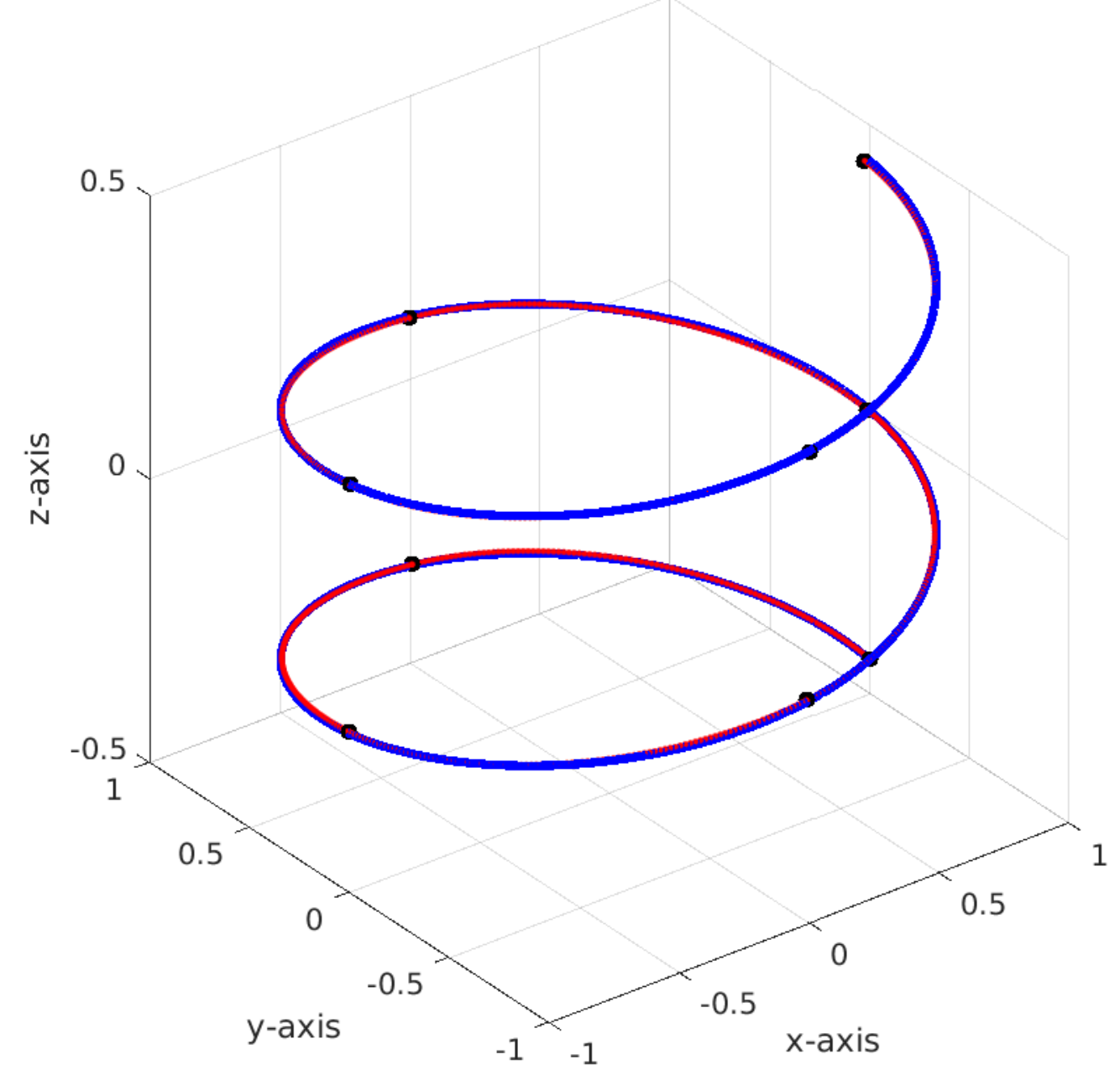}\\
 \includegraphics[width=2in]{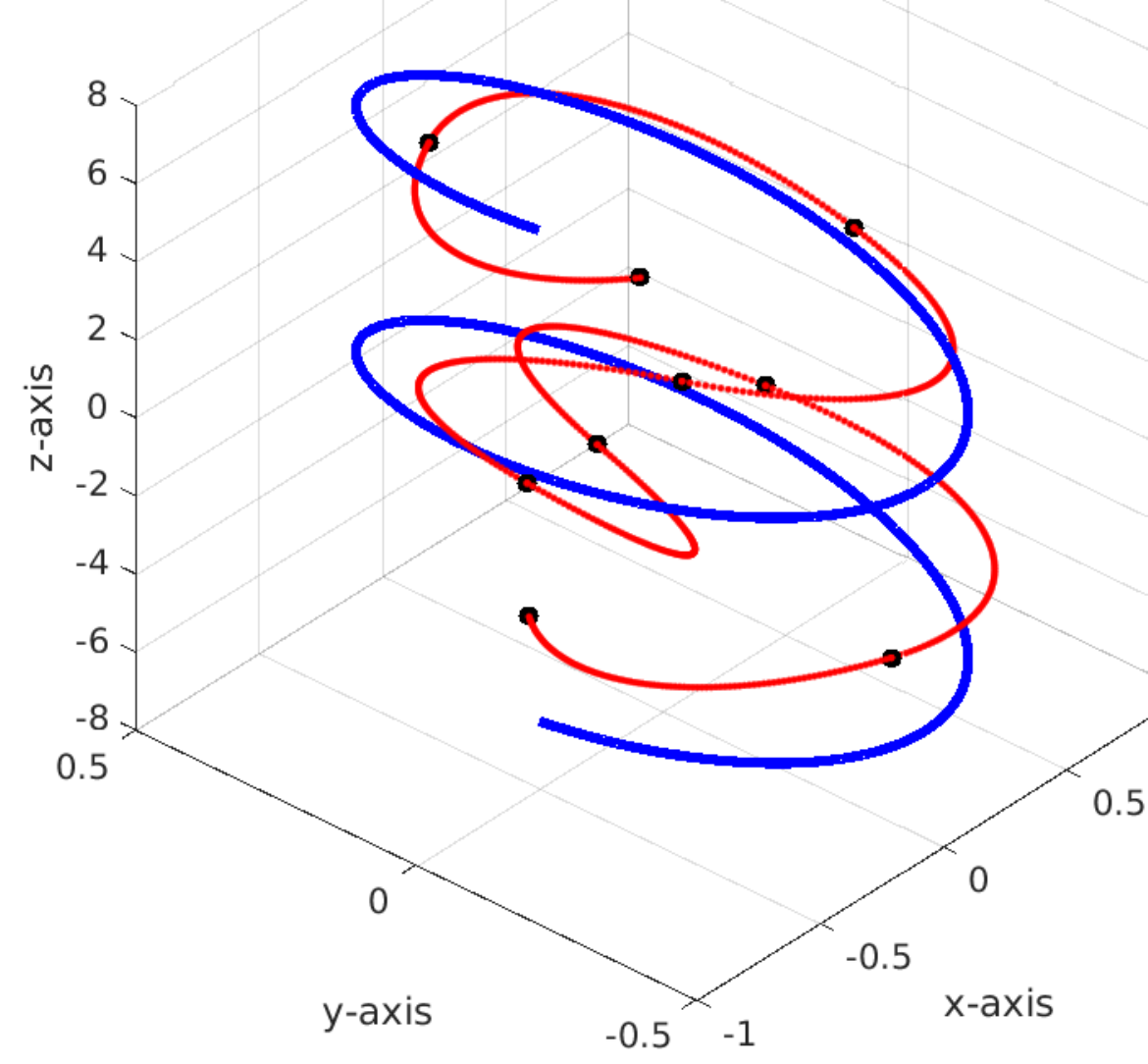} &
 \includegraphics[width=2in]{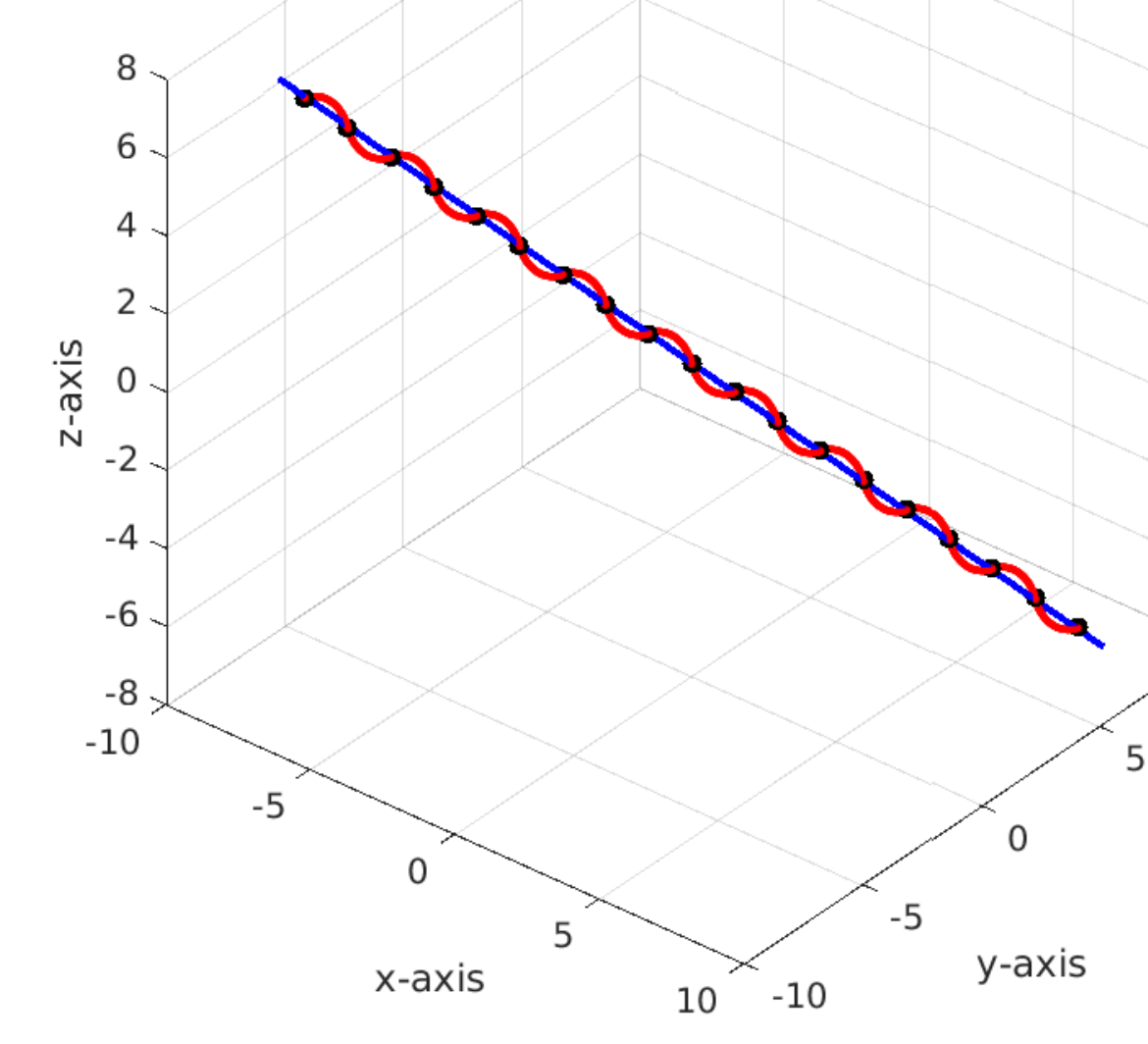}
 \end{array}$
 \end{center}
 \caption{Curve approximation using open tangles: Helices are shown in blue in each plot, while the tangle is shown in red with black points depicting the knots. The parameters $(a,b)$ in Equation \eqref{eqn:helix} are (top-left):$(1.3,0.2)$, (top-right):$(1,0.07)$, (bottom-left):$(-0.5,1)$, while the straight line in the final plot is $h(t) = (-t,-t,t)^T$. The straight line is approximated with an $18$-link open tangle, while the rest are approximated using $8$-link tangles. (top-right) Observe that the tangle almost coincides with the helix of curvature $0.9951$. Note that in all examples, the two curves have been aligned to remove any difference arising out of rigid motions using Procrustes analysis. (bottom-left) Since curvature of a tangle is fixed to $\pm 1$, it can yield more \emph{loops} if used to approximate helices with a low curvature (here $0.4$).}
 \label{fig:OT_curveapp}
 \end{figure*}

One may expect the tangle configurations to be able to replicate helical structures very well. As it turns out, very few helices can be represented by tangles. A helix has the following parameterization for $t \in [0,\frac{n\pi}{2}]$,
\begin{align}
\label{eqn:helix}h(t) = \left(a\cos t,a\sin t,bt\right)^T. 
\end{align}
The tangent vector at any $t \in [0,\frac{n\pi}{2}]$, is
\begin{align*}
\dot{h}(t) = &\left(-a\sin t,a\cos t,b\right)^T. 
\end{align*}
Even thought $h$ is not unit speed, it is a constant speed parameterization. At knot points $t_i = \frac{i\pi}{2}, i=0,\ldots,n$, the tangents are
\begin{align*}
\dot{h}(t_i) = \left(-a\frac{1-(-1)^i}{2},a\frac{1+(-1)^i}{2},b\right)^T. 
\end{align*}
As can be seen from the above equation, in order for a helix to be a tangle curve, $\ip{\dot{h}(t_i)
}{\dot{h}(t_{i+1})} = 0, i = 0,\ldots,n-1$, which implies $b=0$. The same conclusion can be drawn from the fact that curvature of helix at any point is $\frac{a}{\sqrt{a^2+b^2}}$, while that of a tangle is $\pm 1$, wherever it exists. Thus for the curves to match, $b=0$. In Figure\ref{fig:OT_curveapp}, we show results of our curve approximation algorithms for helices with various curvatures: $0.7514, 0.9951, 0.4$, and a straight line, using open tangles.

\begin{figure*}[t]
\begin{center}$
\begin{array}{cc}
\includegraphics[width=2in]{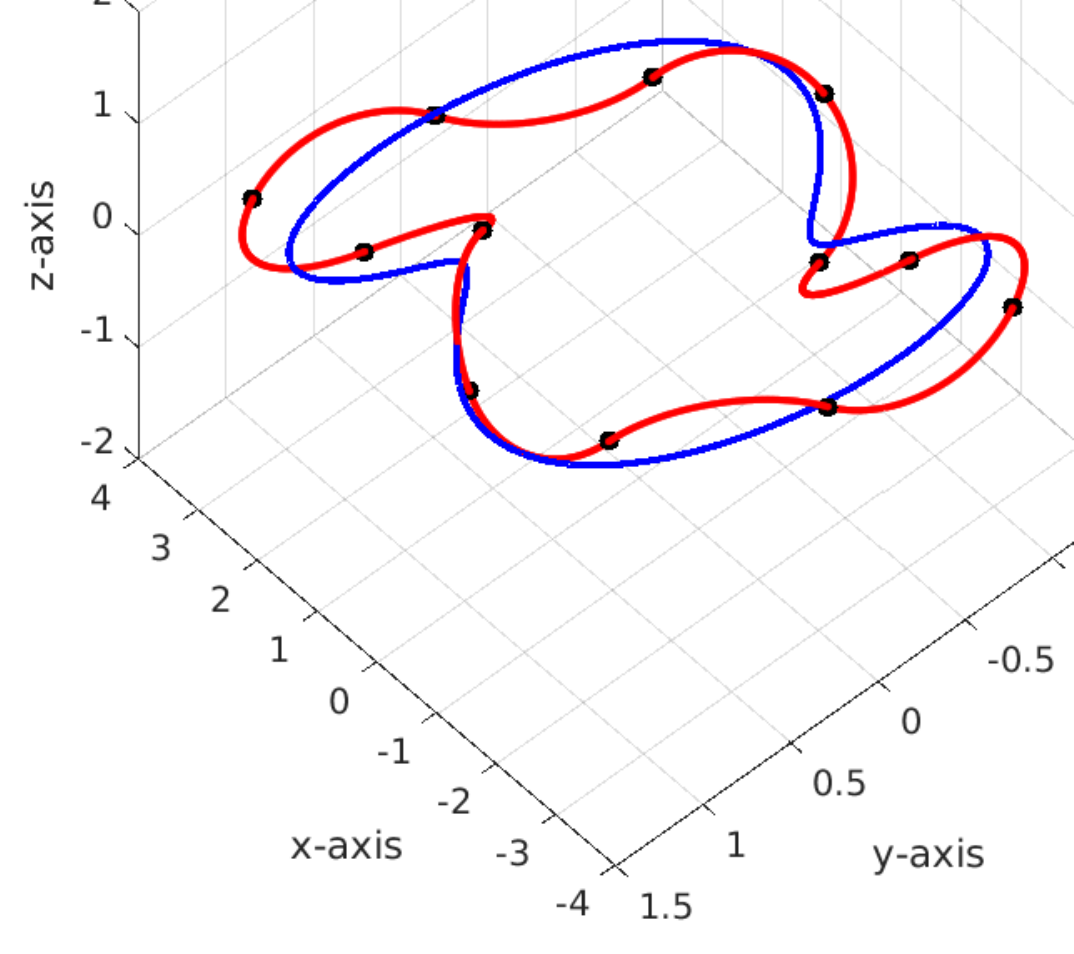} &
\includegraphics[width=2in]{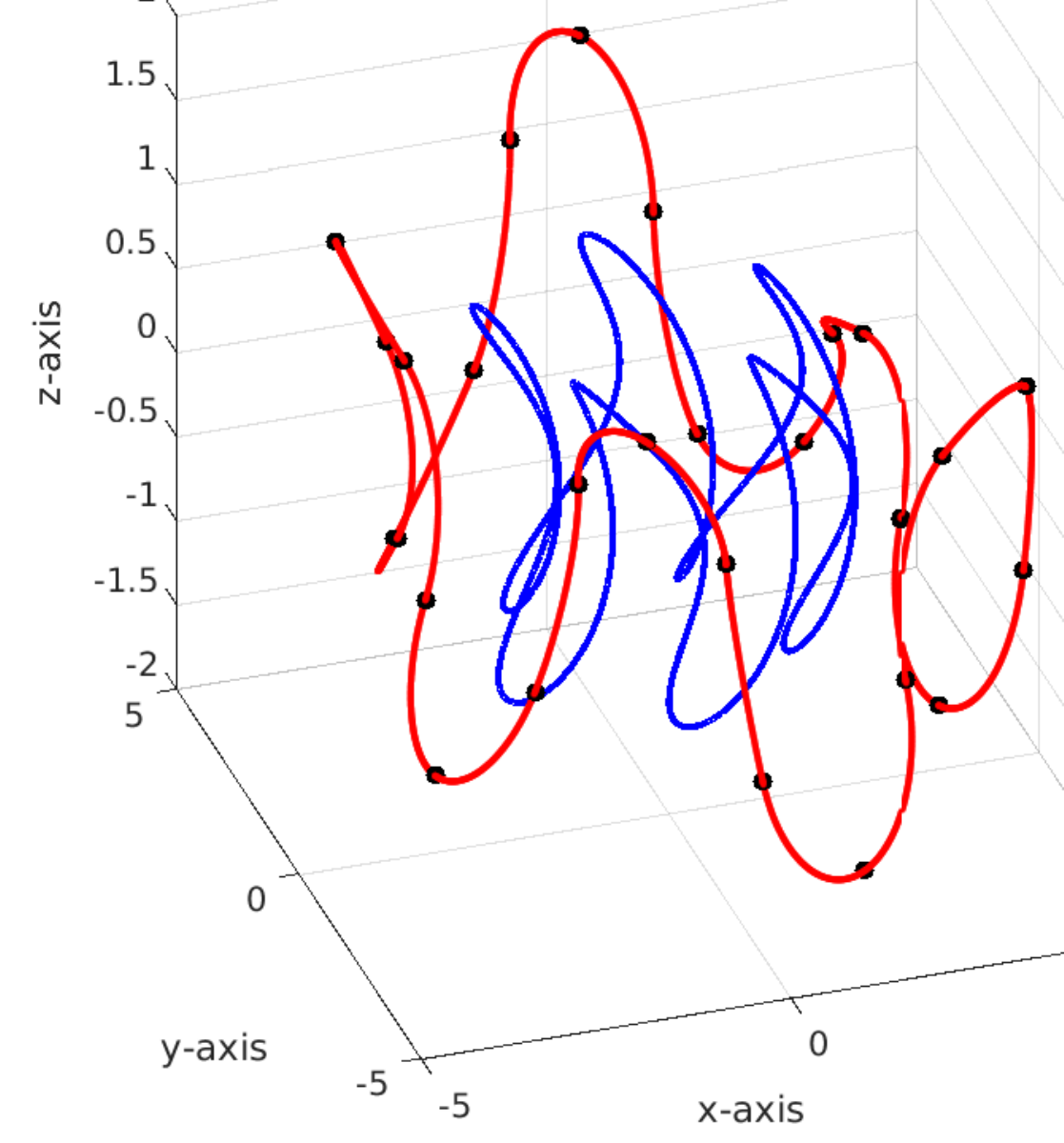}
\end{array}$
\end{center}
\caption{Curve approximation using closed tangles. Curves to be approximated are shown in blue, while tangle curves are shown in red. The parameters $(a,b,c,d)$ used in Equation \eqref{eqn:CTCA} to generate curves shown above are: (left) $(2,2,1,1)$, (right) $(2,5,5,5)$. For the figure on the left, $12$ links have been used, while the figure on the right, $27$ links have been used. Note that in all examples, the two curves have been aligned to remove any difference arising out of rigid motions using Procrustes analysis.}
\label{fig:CT_curveapp}
\end{figure*}

For curve approximation using closed tangles, we consider space curves of the form
\begin{align}
h(t) = & (a\cos t + \cos(bt)\cos(ct), a\sin t + \cos(bt)\sin(ct),\\
       & d\sin(ct))^T, t \in [0,2\pi],
\label{eqn:CTCA}       
\end{align}
where $a,b,c$ and $d$ are real constants. We show couple of examples of such curves and the closed tangles that approximate them in Figure \ref{fig:CT_curveapp}. As can be seen from this analysis, using tangles for curve approximation will find only limited applications, perhaps to model helical structures in protein molecules.
\subsection{Geodesics on Tangle space}
\begin{figure*}[h]
\begin{center}$
\begin{array}{cc}
\includegraphics[width=2in]{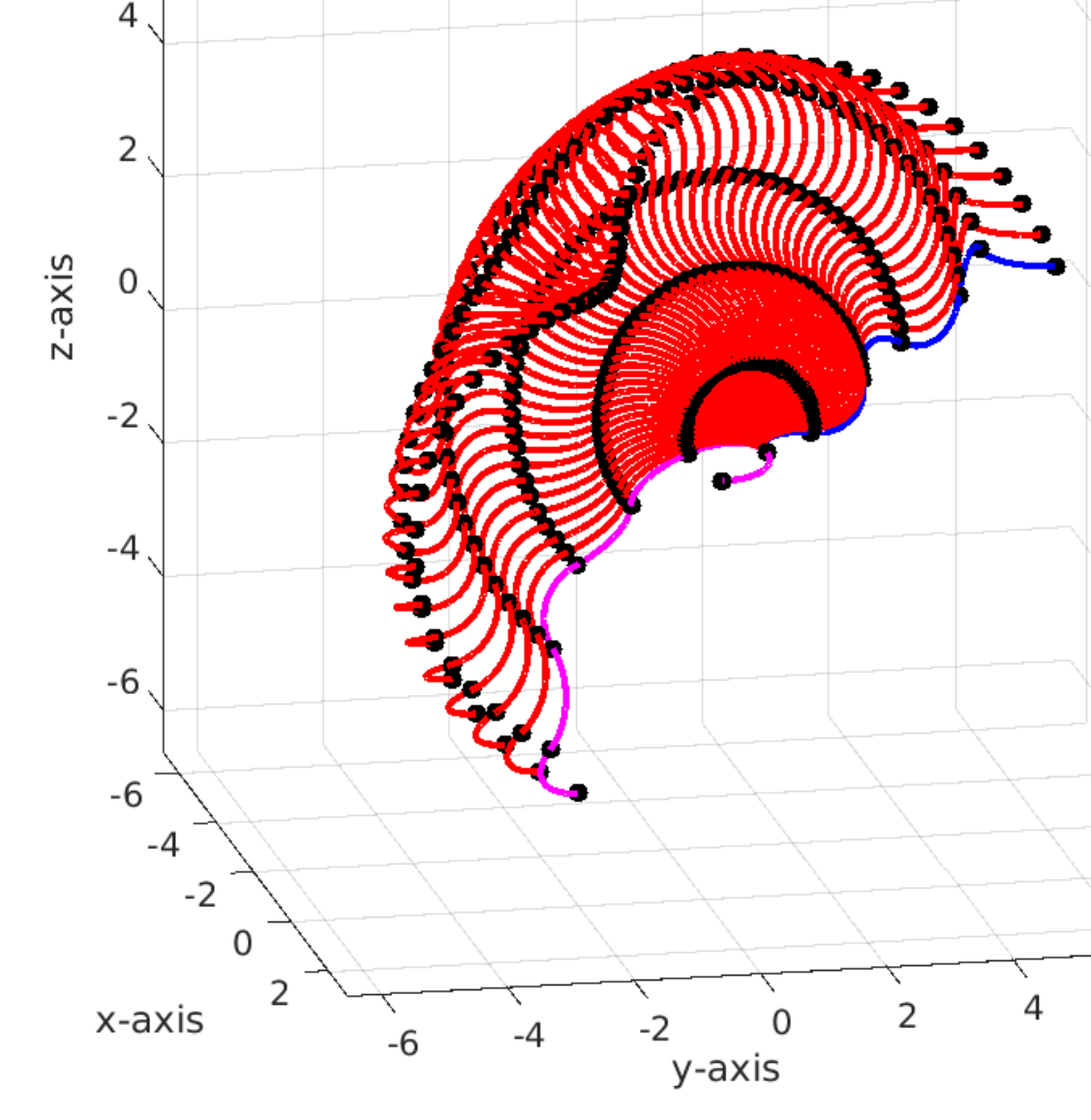} &
\includegraphics[width=2in]{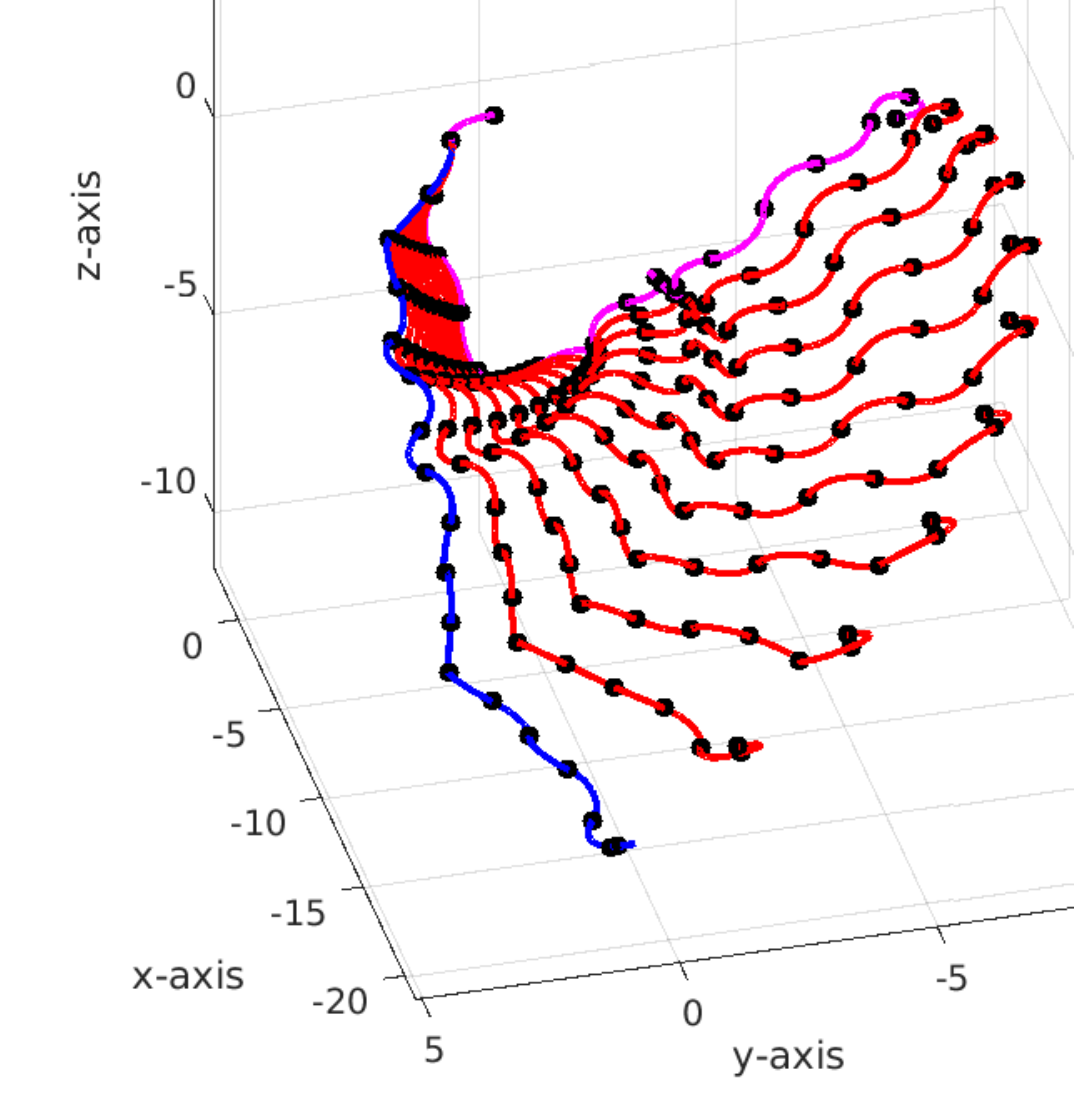}
\end{array}$
\end{center}
\caption{Geodesics on $n$-link open tangle shape space. Initial tangles are shown in blue, final in magenta, and intermediate ones in red. The initial tangles and the tangent vector used in the Exp map to generate the final tangles were chosen randomly. The geodesics shown above may not necessary be length minimizing geodesics. Also note that the Exp map is computed on the shape space of $n$-link open tangles, thus the first link is common to all tangles. This is why the first link is shown only in magenta(final tangle). (left) Tangles with $7$ links, (right) Tangles with $18$ links.}
\label{fig:geoOT}
\end{figure*}

As shown in Section \ref{sec:modeltangle}, geodesics on $n$-link open tangle space can be obtained via the Exp map on $\mathbb{T}^{n-1}$ (refer Equation \eqref{eqn:expso}). For a few $p \in \mathbb{T}^{n-1}$ and $v \in T_{p}(\mathbb{T}^{n-1})$, samples of the geodesic $\gamma:t \mapsto Exp_{p}(tv)$ are shown in Figure \ref{fig:geoOT}. Note here apart from round-off/precision errors, there are no other numerical inaccuracies, since the Exp map has an explicit form.

\begin{figure*}[t]
\begin{center}$
\begin{array}{ccc}
\includegraphics[width=2in]{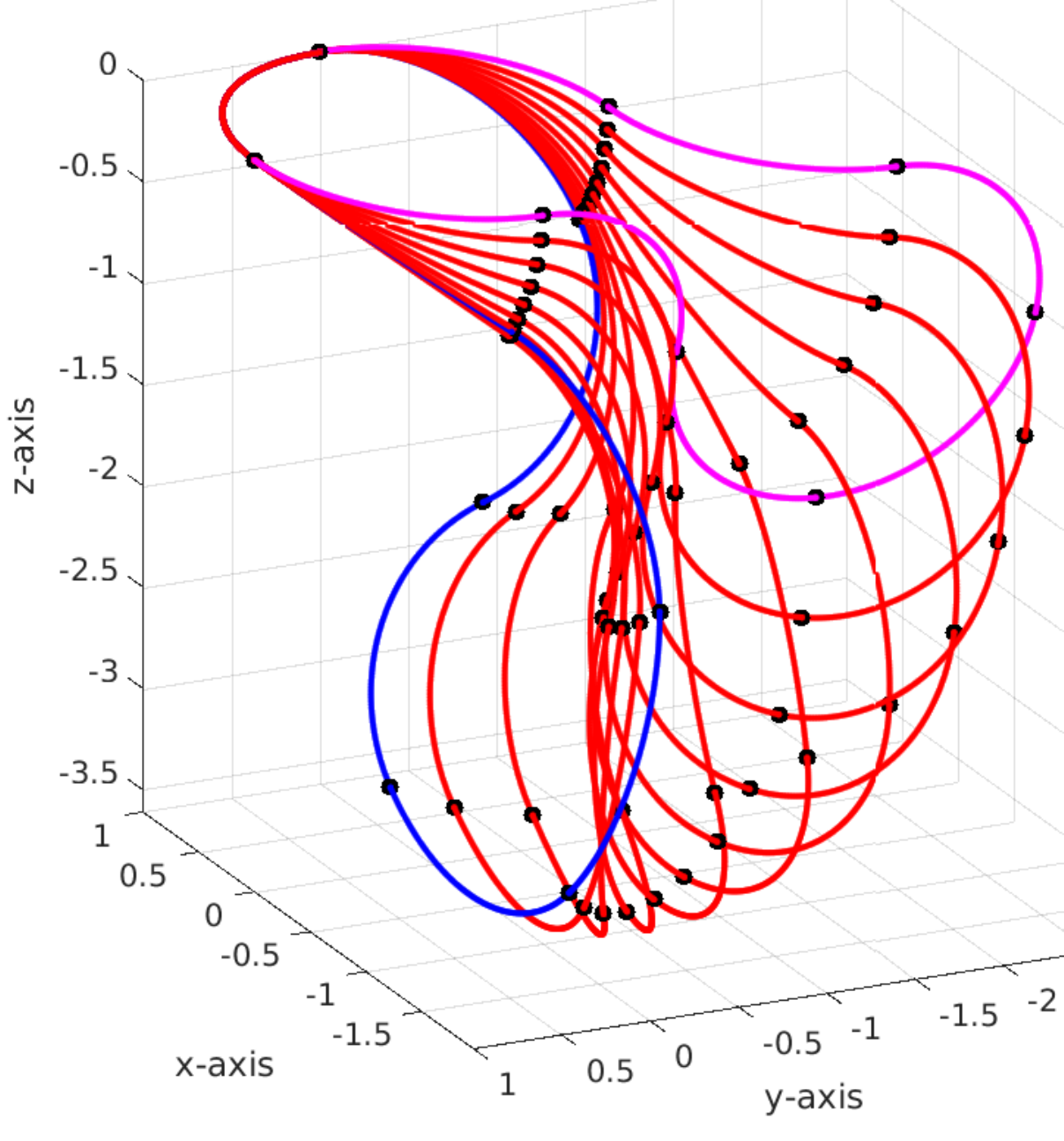} &
\includegraphics[width=2in]{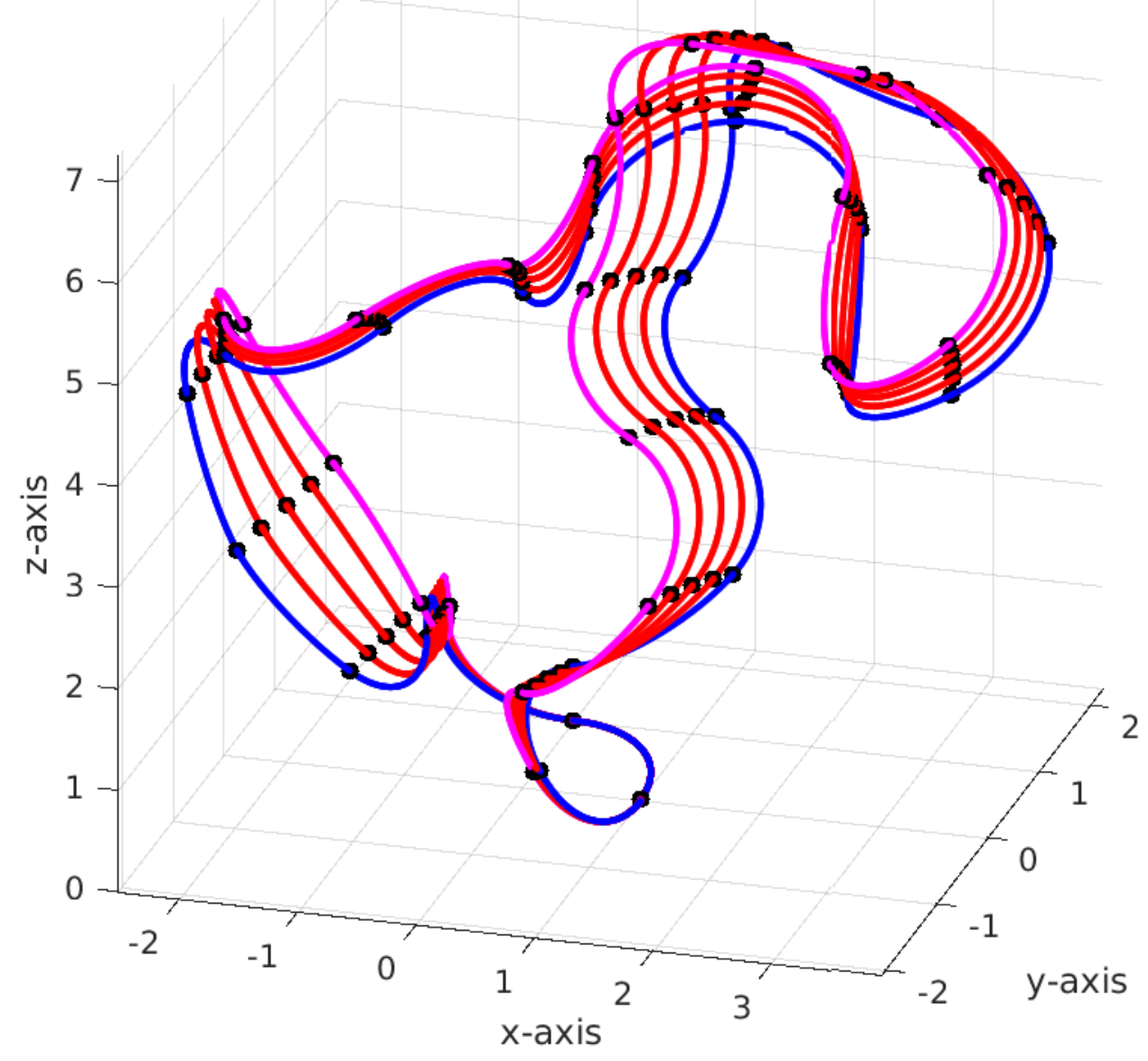} &
\includegraphics[width=2in]{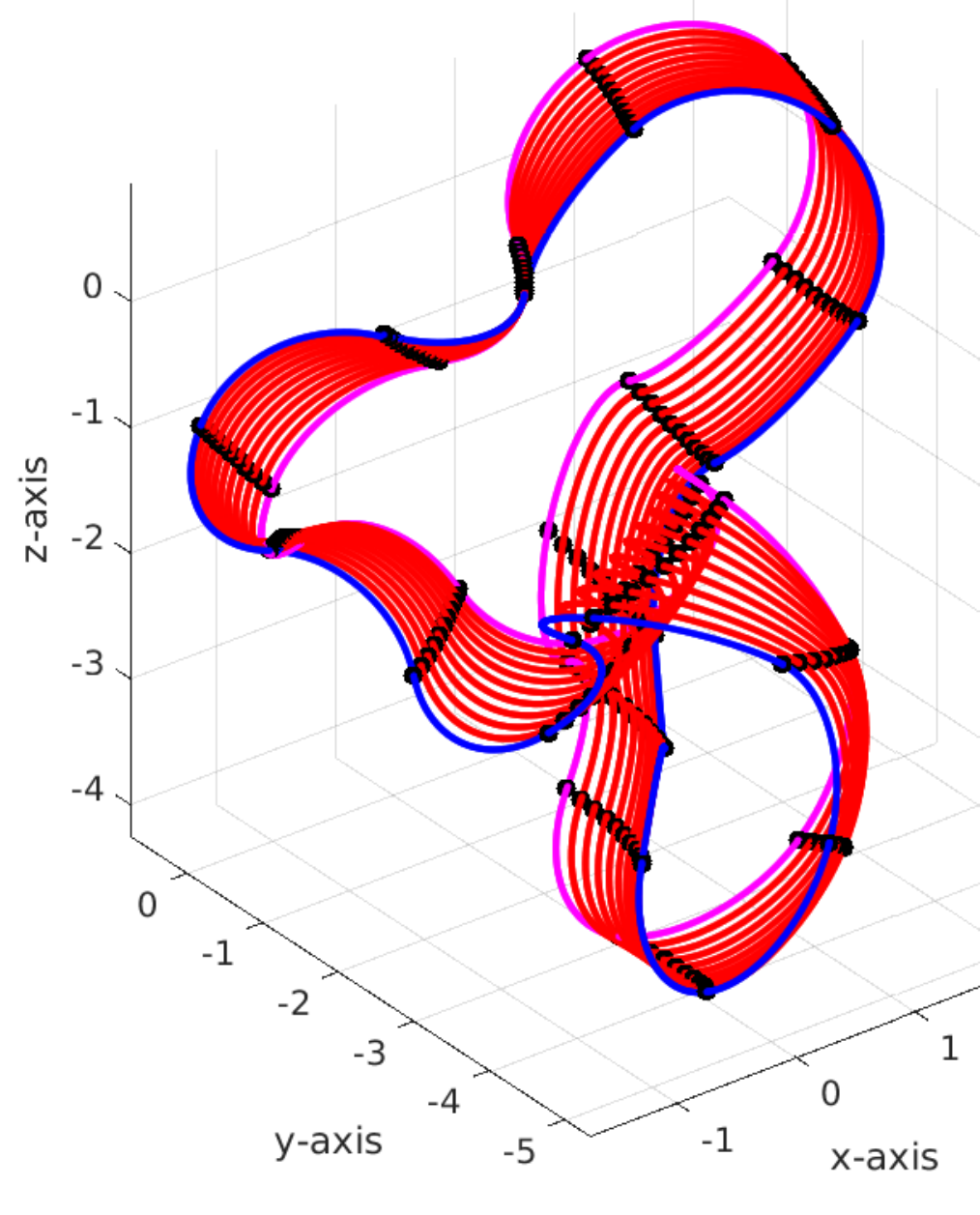} 
\end{array}$
\end{center}
\caption{Geodesics on $n$-link closed Tangle space. While the plot on the left and center depicts geodesics on the tangle shape space, the plot on the right represents geodesics on the tangle space (without restricting the first link). The initial tangles are shown in blue, the final in magenta, and some intermediate ones are shown in red. The number of links is $8$, $24$ and $18$ from left to right. Observe that all tangles in the plot on the left and center, have the first link common (canonical quarter of unit circle).}
\label{fig:geoCT}
\end{figure*}

Geodesics on $n$-link closed tangle space are much harder to compute as shown in Section \ref{sec:comptools}.    Examples of geodesics computed via the Exp map defined earlier, are shown in Figure \ref{fig:geoCT}. We demonstrate two cases here: geodesics on the $n$-link closed tangle shape space $S_c$, and geodesics on the $n$-link closed tangle space. The former being different from the latter in the aspect that global rigid motions have been factored out by fixing the first link to be the canonical quarter of unit circle going from $(0,1,0)$ to $(1,0,0)$. In all examples shown in Figure \ref{fig:geoCT}, the maximum deviation of angles between tangent vectors of consecutive links from right angles is strictly less than $0.1$ degrees, while all tangent vectors computed are vectors with a maximum deviation less than $0.01$ from being unit norm. This shows the accuracy of the computational method.
\subsection{Log map}
Log map on $n$-link open tangle shape space has the closed form description given in Equation \eqref{eqn:logso}. We do not show results for the Log map on open tangle shape space, since both Exp map and Log map have nice closed form solutions, and we observe $Exp_p(Log_p(q)) = q$ upto round-off/precision errors. Let $v \in T_p(S_o)$ be such that $q = Exp_p(v)$, then $Log_p(q)$ may not equal $v$ depending on whether $q$ is in the cut-locus of $p$.   

\begin{figure*}[h]
 \begin{center}$
 \begin{array}{cc}
 \includegraphics[width=2in]{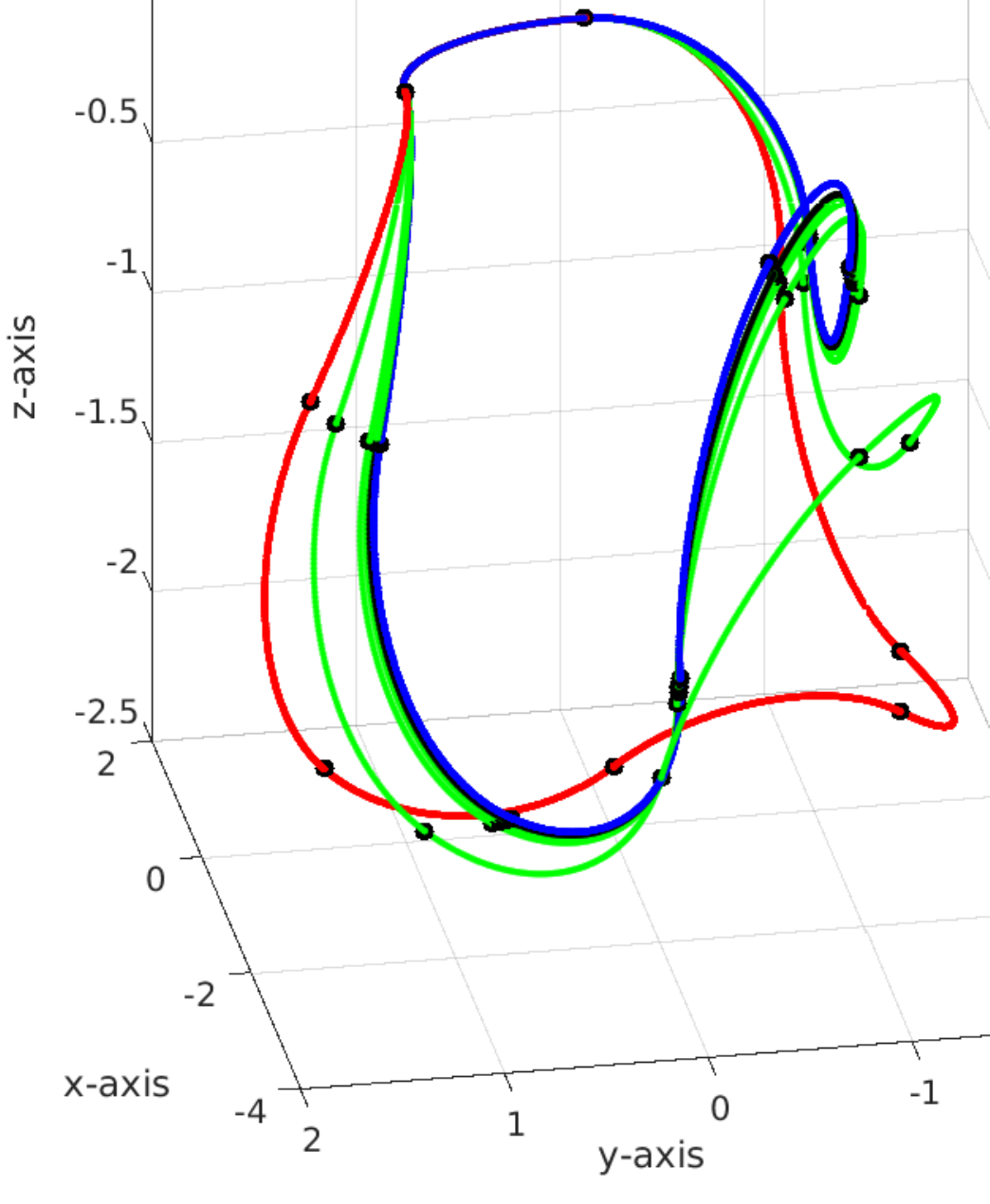} &
 \includegraphics[width=2.5in]{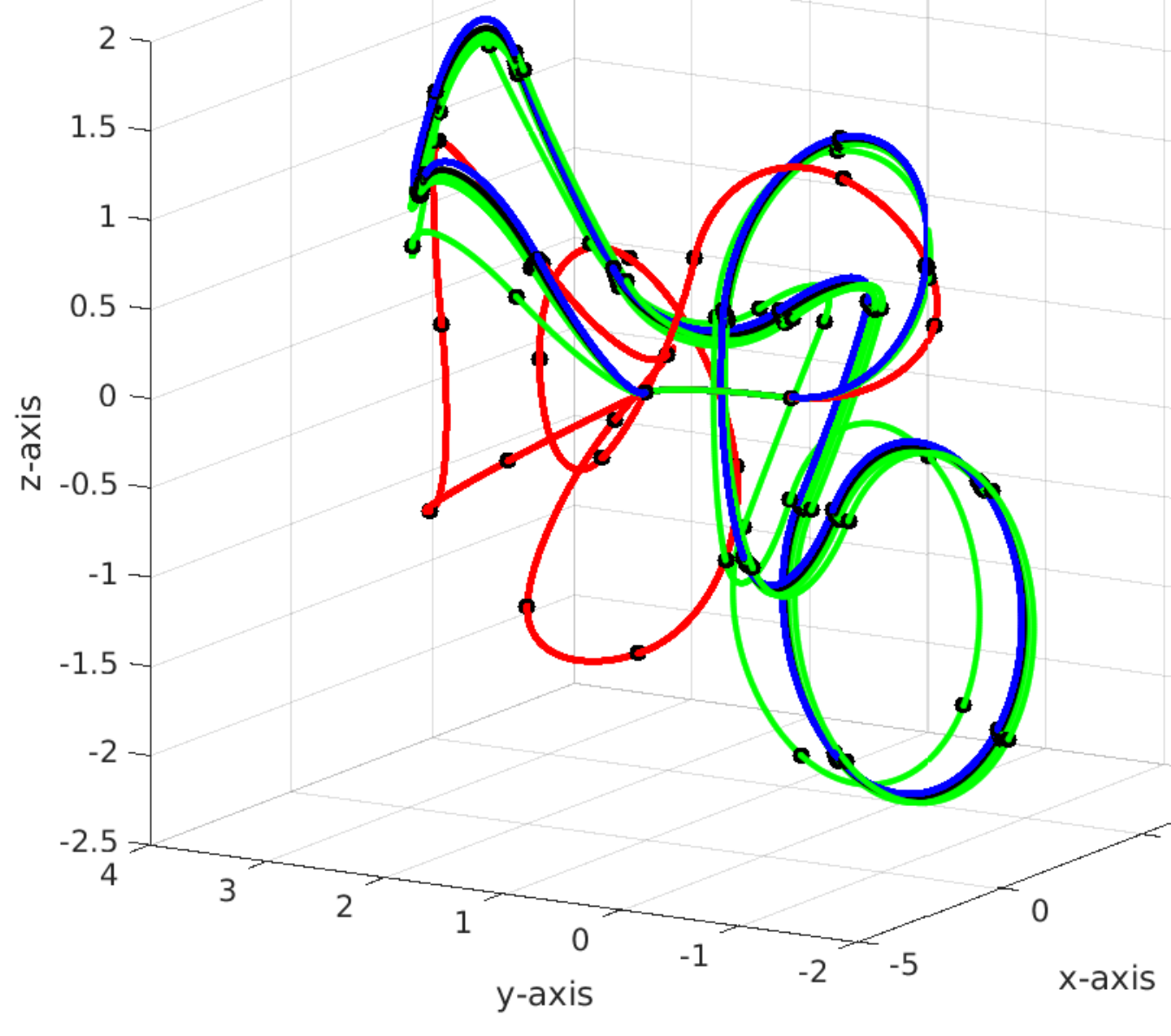}
 \end{array}$
 \end{center}
 \caption{Log Map on $n$-link closed tangle shape space. The tangles in red are the points where the Log map is to be computed, while the tangles in blue are the points to which the Log map is to be computed. The tangles in black are the exponential map of the computed Log map, while tangles in green show the exponential map of the intermediate estimates of the Log map. The number of links is $8$ for the plot on the left and $18$ for the plot on the right. }
 \label{fig:Logmap}
 \end{figure*}

Results for Log map on the $n$-link closed tangle shape space are shown in Figure \ref{fig:Logmap}. The algorithm stops when the Euclidean distance between the target tangle curve and the exponential of the estimated Log map is below a certain threshold. Note that for any tangle obtained while computing the Log map, the maximum deviation of angles between tangent vectors of consecutive links from right angles was strictly less than $0.1$ degrees, while all tangent vectors computed were vectors with a maximum deviation less than $0.01$ from being unit norm.
\section{Discussion and Conclusion}
\label{sec:discon}
\begin{figure*}[h]
\begin{center}$
\begin{array}{cc}
\includegraphics[width=2in]{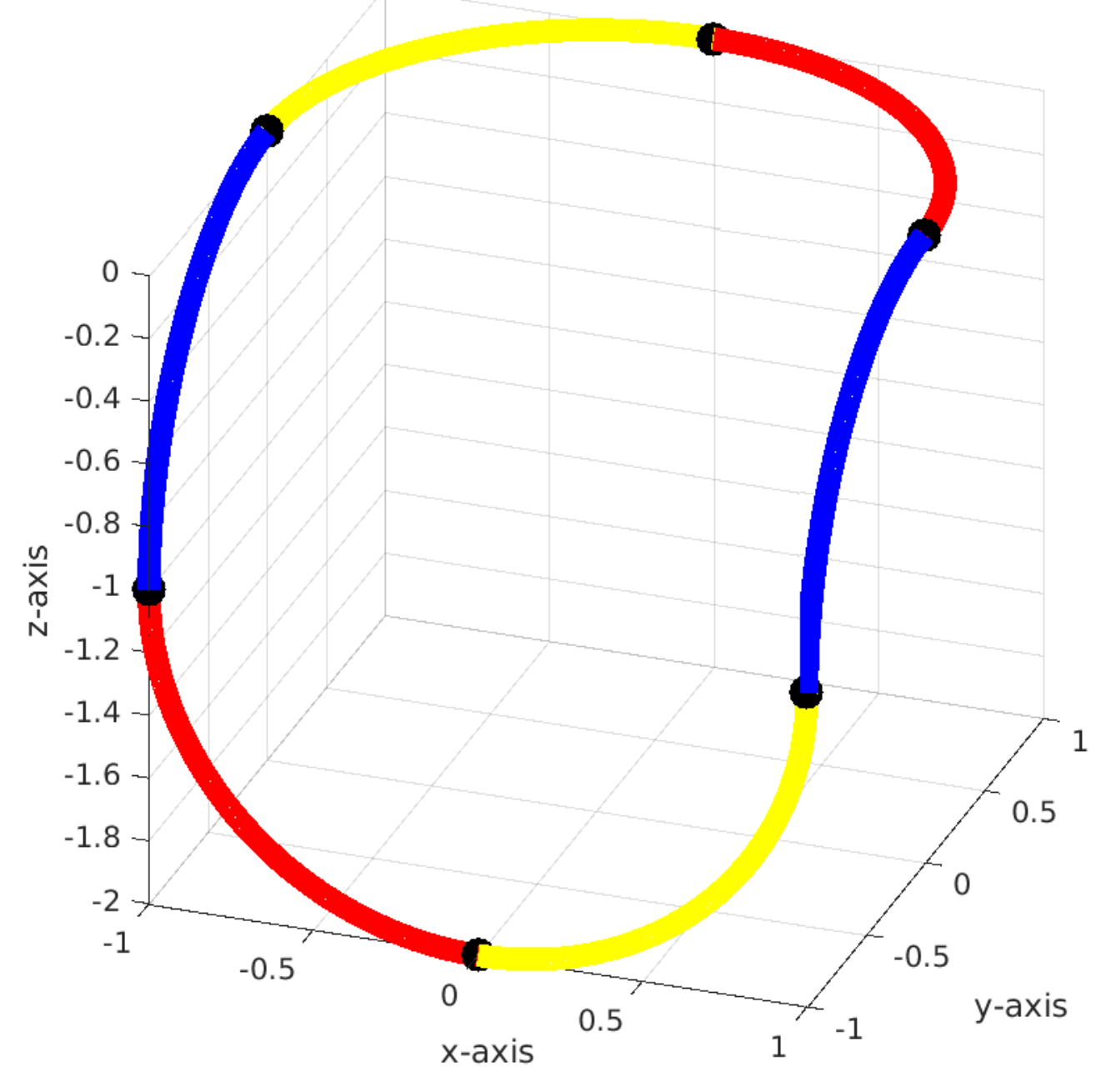} &
\includegraphics[width=2in]{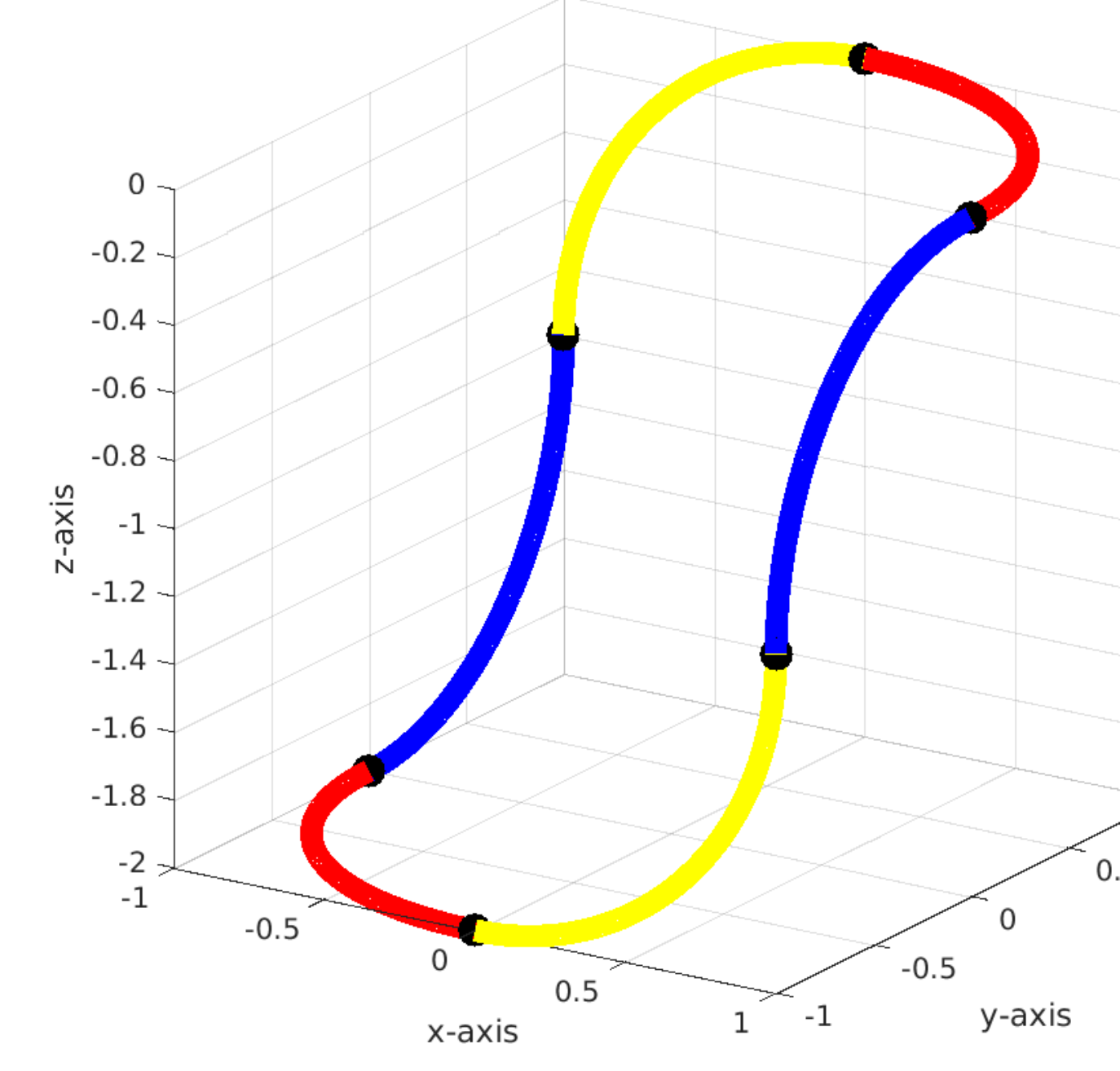}
\end{array}$
\end{center}
\caption{Two $6$-link closed tangle shapes. Apart from the first two tangent vectors being set to $(0,1,0)^T$ and $(-1,0,0)^T$ for both examples, the rest four tangent vectors are: (left) - $(0,-1,0)^T,(0,0,-1)^T,(1,0,0)^T,(0,0,1)^T$, (right) $(0,0,-1)^T,(0,-1,0)^T,(1,0,0)^T,(0,0,1)^T$. }
\label{fig:cntexample}
\end{figure*}

\begin{figure*}[h]
\centering
\includegraphics[width=2in]{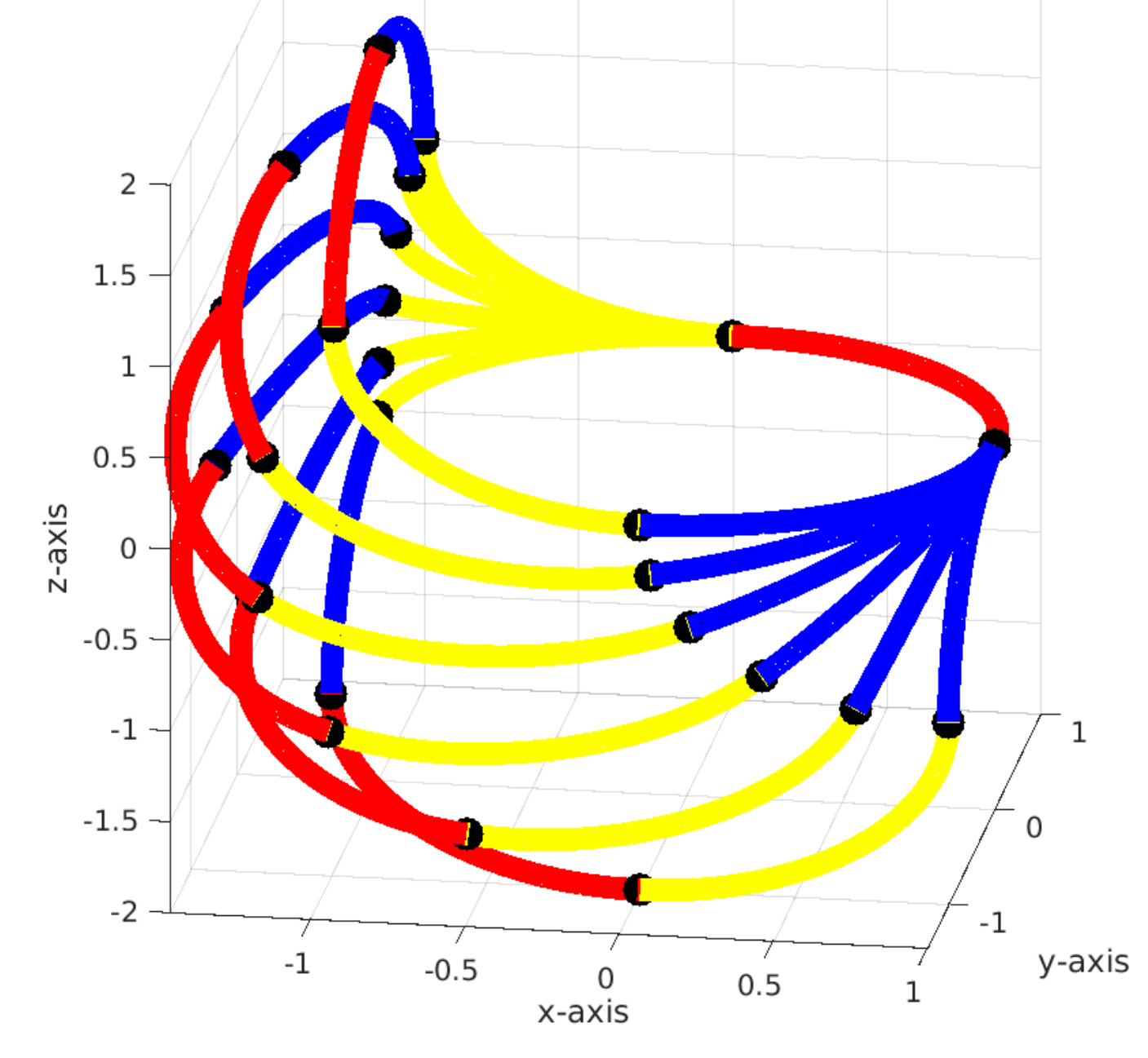}
\caption{Samples of geodesic along the only tangent vector for the tangle shown in the left in Figure\ref{fig:cntexample}. As mentioned in a footnote in Section\ref{sec:modeltangle}, even after fixing the first link, we may obtain different tangles with the same shape due to additional symmetries (which as of now are ignored), as can be seen above.}
\label{fig:Exptangle6}
\end{figure*}

We now address the question: Is the $n$-link closed tangle shape space a manifold or not. In case $S_c = F^{-1}(0)$ is a manifold, by the constant rank level set theorem\cite{Lee2003}, the rank of $F$ needs to be constant on $S_c$. The tangles provided in Figure \ref{fig:cntexample} give a counterexample. For $6$-link closed tangles, the constraint function defined after Equation \eqref{eqn:Sc} is of the kind $F:\Rr^{12} \rightarrow \Rr^{12}$. As it turns out, for the tangle shown on the left in Figure\ref{fig:cntexample}, the rank of $F$ (defined as rank of $JF$) is $11$, while the rank of $F$ for the tangle shown on the right is $12$. This makes the Jacobian a full rank matrix. Thus there is neighborhood in which the Jacobian remains full rank, implying that the two tangle configurations belong to two different connected components of $S_c$. The tangle on the left is part of a $1$D path on $S_c$ ($1$ D tangent space), while the latter is an isolated point ($0$ D tangent space). This confirms that $S_c$ with $6$ links is not a manifold and contains at least two connected components, one being a $1$D path (in fact a closed one), while the other being an isolated point. Few samples of the geodesic for the tangle on the left from Figure \ref{fig:cntexample} along the only tangent vector is shown in Figure \ref{fig:Exptangle6}. When the norm of the tangent vector is increased, the corresponding geodesic reaches the initial tangle, thus forming a closed path in $S_c$. 
Having said that, we conjecture that with a larger $n$, $S_c$ has only one connected component. This stems from the intuition that with more degrees of freedom available, it should be possible to deform one tangle configuration to another.

Forming complex space curves from arcs of planar circles is a fascinating aspect of tangles. This is analogous to creating $3$D meshes from planar triangles. It would be interesting to analyze $3$D meshes, where every triangle is obtained via a rigid transformation of a single fixed equilateral triangle. The expressibility of such a mesh would be limited though.

To conclude, we summarize our contribution. We have given two models of $n$-link open and closed tangle shape spaces. We have shown that tangles are a subset of trigonometric splines. We have presented algorithms for space curve approximation, computing geodesics and Log map on the space of $n$ link closed/open tangles. 
% or


%\appendices
%% you can choose not to have a title for an appendix
%% if you want by leaving the argument blank
%\section{}
%% use section* for acknowledgment
%\section*{Acknowledgment}
%% Can use something like this to put references on a page
%% by themselves when using endfloat and the captionsoff option.
%\ifCLASSOPTIONcaptionsoff
%  \newpage
%\fi
%
%
%
%% trigger a \newpage just before the given reference
%% number - used to balance the columns on the last page
%% adjust value as needed - may need to be readjusted if
%% the document is modified later
%%\IEEEtriggeratref{8}
%% The "triggered" command can be changed if desired:
%%\IEEEtriggercmd{\enlargethispage{-5in}}
%
%% references section
%
%% can use a bibliography generated by BibTeX as a .bbl file
%% BibTeX documentation can be easily obtained at:
%% http://mirror.ctan.org/biblio/bibtex/contrib/doc/
%% The IEEEtran BibTeX style support page is at:
%% http://www.michaelshell.org/tex/ieeetran/bibtex/
\bibliographystyle{IEEEtran}
%% argument is your BibTeX string definitions and bibliography database(s)
\bibliography{tanglebib}

\begin{thebibliography}{10}
\providecommand{\url}[1]{#1}
\csname url@samestyle\endcsname
\providecommand{\newblock}{\relax}
\providecommand{\bibinfo}[2]{#2}
\providecommand{\BIBentrySTDinterwordspacing}{\spaceskip=0pt\relax}
\providecommand{\BIBentryALTinterwordstretchfactor}{4}
\providecommand{\BIBentryALTinterwordspacing}{\spaceskip=\fontdimen2\font plus
\BIBentryALTinterwordstretchfactor\fontdimen3\font minus
  \fontdimen4\font\relax}
\providecommand{\BIBforeignlanguage}[2]{{%
\expandafter\ifx\csname l@#1\endcsname\relax
\typeout{** WARNING: IEEEtran.bst: No hyphenation pattern has been}%
\typeout{** loaded for the language `#1'. Using the pattern for}%
\typeout{** the default language instead.}%
\else
\language=\csname l@#1\endcsname
\fi
#2}}
\providecommand{\BIBdecl}{\relax}
\BIBdecl

\bibitem{Kendall1989}
D.~G. Kendall, ``A survey of the statistical theory of shape,''
  \emph{Statistical Science}, vol.~4, pp. 87 -- 99, 1989.

\bibitem{Thompson1917}
D.~W. Thompson, \emph{On Growth and Form}, 1917.

\bibitem{Kendall1984}
D.~G. Kendall, ``Shape manifolds, procrustean metrics, and complex projective
  spaces,'' \emph{Bulletin of the London Mathematical Society}, vol.~16, pp.
  81--121, 1984.

\bibitem{Bookstein1986}
F.~L. Bookstein, ``Size and shape spaces for landmark data in two dimensions,''
  \emph{Statistical Science}, vol.~1, no.~2, pp. 181--222, 1986.

\bibitem{Michor2007}
P.~W. Michor and D.~Mumford, ``An overview of the riemannian metrics on spaces
  of curves using the hamiltonian approach,'' \emph{Applied and Computational
  Harmonic Analysis}, vol.~23, no.~1, pp. 74--113, July 2007.

\bibitem{Younes1998}
L.~Younes, ``Computable elastic distances between shapes,'' \emph{SIAM Journal
  on Applied Mathematics}, vol.~58, pp. 565 -- 586, 1998.

\bibitem{Klassen2004}
E.~Klassen, A.~Srivastava, W.~Mio, and S.~Joshi, ``Analysis of planar shapes
  using geodesic paths on shape spaces,'' \emph{IEEE Transactions on Pattern
  Analysis and Machine Intelligence}, vol.~26, pp. 372 -- 383, 2004.

\bibitem{Blum1973}
H.~Blum, ``Biological shape and visual science,'' \emph{Journal of Theoretical
  Biology}, vol.~38, pp. 205--287, 1973.

\bibitem{Giblin1999}
P.~J. Giblin and B.~B. Kimia, ``On the local forms and transition of symmetry
  sets, medial axis and shocks in 2d,'' \emph{International Conference on
  Computer Vision}, pp. 385--391, 1999.

\bibitem{Sebastian2001}
T.~B. Sebastian, P.~N. Klein, and B.~B. Kimia, ``Recognition of shape by
  editing shock graphs,'' \emph{International Conference on Computer Vision},
  pp. 755--762, 2001.

\bibitem{Sommer2009}
S.~Sommer, A.~Tatu, C.~Chen, M.~de~Bruijne, M.~Loog, D.~Jorgensen, M.~Nielsen,
  and F.~Lauze, ``Bicycle chain shape models,'' \emph{CVPR Workshop on
  Mathematical Methods in Biomedical Image Analysis}, 2009.

\bibitem{Glaunes2006}
J.~Glaun{\`e}s, A.~Trouv{\'e}, and L.~Younes, \emph{Statistics and Analysis of
  Shapes}.\hskip 1em plus 0.5em minus 0.4em\relax Birkh\"auser Boston, 2006,
  ch. Modeling planar shape variation via Hamiltonian flows of curves, pp.
  335--361.

\bibitem{Srivastava2011}
\BIBentryALTinterwordspacing
A.~Srivastava, E.~Klassen, S.~H. Joshi, and I.~H. Jermyn, ``Shape analysis of
  elastic curves in euclidean spaces,'' \emph{IEEE Trans. Pattern Anal. Mach.
  Intell.}, vol.~33, no.~7, pp. 1415--1428, Jul. 2011. [Online]. Available:
  \url{http://dx.doi.org/10.1109/TPAMI.2010.184}
\BIBentrySTDinterwordspacing

\bibitem{Kessler2007}
P.~Kessler, ``On the shapes of tangles curves,'' Ph.D. dissertation, University
  of California Berkeley, 2007.

\bibitem{Reti1993}
T.~R\'eti and I.~Czinege, ``Generalized fourier descriptors for shape analysis
  of 3-dimensional closed curves,'' \emph{Acta Stereologica}, vol.~12, no.~2,
  pp. 95--102, December 1993.

\bibitem{Schoenberg1964}
\BIBentryALTinterwordspacing
I.~J. Schoenberg, ``Spline interpolation and best quadrature formulae,''
  \emph{Bull. Amer. Math. Soc.}, vol.~70, no.~1, pp. 143--148, 01 1964.
  [Online]. Available: \url{http://projecteuclid.org/euclid.bams/1183525790}
\BIBentrySTDinterwordspacing

\bibitem{Boor1962}
C.~De~Boor, ``Bicubic spline interpolation,'' \emph{Journal of mathematics and
  physics}, vol.~41, no.~1, pp. 212--218, 1962.

\bibitem{Unser1993}
M.~Unser, A.~Aldroubi, and M.~Eden, ``B-spline signal processing: Part 1 -
  theory,'' \emph{IEEE Transactions on Signal Processing}, vol.~41, no.~2, pp.
  821--832, February 1993.

\bibitem{Bartels1987}
R.~H. Bartels, J.~C. Beatty, and B.~A. Barsky, \emph{An Introduction to Splines
  for Use in Computer Graphics \& Geometric Modeling}.\hskip 1em plus 0.5em
  minus 0.4em\relax San Francisco, CA, USA: Morgan Kaufmann Publishers Inc.,
  1987.

\bibitem{Weiss1997}
\BIBentryALTinterwordspacing
V.~Wei{\ss}, H.-P. Seidel, G.~Greiner, T.~Puchtler, and W.~Eberlein,
  \emph{Optimizing CNC Programs Using Spline Techniques}.\hskip 1em plus 0.5em
  minus 0.4em\relax Berlin, Heidelberg: Springer Berlin Heidelberg, 1997, pp.
  70--82. [Online]. Available:
  \url{http://dx.doi.org/10.1007/978-3-642-60607-6_6}
\BIBentrySTDinterwordspacing

\bibitem{Schumaker2007}
L.~Schumaker, \emph{Spline Functions: Basic Theory}.\hskip 1em plus 0.5em minus
  0.4em\relax Cambridge University Press, 2007.

\bibitem{Champion2009}
\BIBentryALTinterwordspacing
R.~Champion, C.~T. Lenard, and T.~M. Mills, ``A variational approach to
  splines,'' \emph{The ANZIAM Journal}, vol.~42, no.~1, p. 119–135, Feb 2009.
  [Online]. Available:
  \url{https://www.cambridge.org/core/article/a-variational-approach-to-splines/21DBC999A17F5F68C6587000322EEFF5}
\BIBentrySTDinterwordspacing

\bibitem{Wang2002}
H.~Wang, J.~Kearney, and K.~Atkinson, ``Arc-length parameterized spline curves
  for real-time simulation,'' in \emph{In Proc. 5th International Conference on
  Curves and Surfaces}, 2002, pp. 387--396.

\bibitem{Luenberger2015}
D.~G. Luenberger and Y.~Ye, \emph{Linear and Nonlinear Programming}.\hskip 1em
  plus 0.5em minus 0.4em\relax Springer Publishing Company, Incorporated, 2015.

\bibitem{Bertsekas2009}
D.~P. Bertsekas, \emph{Convex optimization theory}, ser. Athena Scientific
  optimization and computation series.\hskip 1em plus 0.5em minus 0.4em\relax
  Belmont, Mass. Athena Scientific, 2009.

\bibitem{Boothby2003}
W.~M. Boothby, \emph{An {I}ntroduction to {D}ifferentiable {M}anifolds and
  {R}iemannian {G}eometry. {R}evised {S}econd {E}dition}.\hskip 1em plus 0.5em
  minus 0.4em\relax Academic Press, 2003.

\bibitem{Dedieu2005}
J.-P. Dedieu and D.~Nowicki, ``Symplectic methods for the approximation of the
  exponential map and the newton iteration on riemannian submanifolds,''
  \emph{Journal of Complexity}, vol.~21, pp. 487 -- 501, 2005.

\bibitem{Mio2004}
W.~Mio and A.~Srivastava, ``Elastic-string models for representation and
  analysis of planar shapes,'' in \emph{Computer Vision and Pattern
  Recognition, 2004. CVPR 2004. Proceedings of the 2004 IEEE Computer Society
  Conference on}, vol.~2.\hskip 1em plus 0.5em minus 0.4em\relax IEEE, 2004,
  pp. II--10.

\bibitem{Noakes1998}
\BIBentryALTinterwordspacing
L.~Noakes, ``A global algorithm for geodesics,'' \emph{Journal of the
  Australian Mathematical Society}, vol.~64, pp. 37--50, 1998. [Online].
  Available: \url{http://www.austms.org.au/Publ/JAustMS/V65P1/abs/n11/}
\BIBentrySTDinterwordspacing

\bibitem{Lee2003}
J.~M. Lee, \emph{Introduction to smooth manifolds}, ser. Graduate texts in
  mathematics.\hskip 1em plus 0.5em minus 0.4em\relax New York, Berlin,
  Heidelberg: Springer, 2003, autre tirage : 2006.

\end{thebibliography}
\end{document}